\documentclass[lettersize,journal]{IEEEtran}
\usepackage{amsmath,amsfonts}
\usepackage{algorithmic}
\usepackage{algorithm}
\usepackage{array}
\usepackage[caption=false,font=normalsize,labelfont=sf,textfont=sf]{subfig}
\usepackage{textcomp}
\usepackage{stfloats}
\usepackage{url}
\usepackage{verbatim}
\usepackage{graphicx}
\usepackage{cite}
\usepackage[dvipsnames]{xcolor}
\usepackage{multirow}
\usepackage{bbding}
\usepackage{soul}
\usepackage{amssymb}
\usepackage{booktabs}
\usepackage[export]{adjustbox}

\usepackage{subfig} 
\usepackage{subfloat}
\usepackage{bm}
\usepackage{hyperref}
\usepackage{flushend}

\begin{document}

\title{Augmentation Matters: A Mix-Paste Method for X-Ray Prohibited Item Detection under Noisy Annotations}

\author{Ruikang Chen, Yan Yan, Jing-Hao Xue, Yang Lu, Hanzi Wang

\thanks{This work
was partly supported by the National Natural Science Foundation of China
under Grants 62372388, 62071404, and U21A20514, and by the
Fundamental Research Funds for the Central Universities under Grant 20720240076. \textit{(Corresponding author: Yan Yan.)}}       
\thanks{R. Chen, Y.~Yan, Y.~Lu, and H.~Wang are with the Fujian Key Laboratory of Sensing and Computing
for Smart City, School of Informatics, Xiamen University,
Xiamen 361005, China
(e-mail: 23020221154074@stu.xmu.edu.cn; yanyan@xmu.edu.cn; luyang@xmu.edu.cn; hanzi.wang@xmu.edu.cn).}
\thanks{J.-H. Xue is with the Department of Statistical Science, University College London, London WC1E 6BT, UK (e-mail: jinghao.xue@ucl.ac.uk).}}

\maketitle

\begin{abstract}
  Automatic X-ray prohibited item detection is vital for public safety. 
  Existing deep learning-based methods all assume that the annotations of training X-ray images are correct.
  However, obtaining correct annotations is extremely hard if not impossible for large-scale X-ray images, where item overlapping is ubiquitous. {As a result, X-ray images are easily contaminated with noisy annotations, leading to performance deterioration of existing methods.} 
  In this paper, we address the challenging problem of training a robust prohibited item detector under noisy annotations (including both category noise and bounding box noise) from a novel perspective of data augmentation, 
 and propose an effective label-aware mixed patch paste augmentation method (\textbf{Mix-Paste}). Specifically, for each item patch, we mix several item patches with the same category label from different images and replace the original patch in the image with the mixed patch. In
  this way, the probability of containing the
  correct prohibited item within the generated image is increased. Meanwhile, the mixing process mimics item overlapping, enabling the model to learn the characteristics of X-ray images.
  Moreover, we design an item-based large-loss suppression (LLS) strategy
  to suppress the large losses corresponding to potentially positive predictions of additional items due to the mixing operation.
  We show the superiority of our method on X-ray datasets under noisy annotations. 
  In addition, we evaluate our method on the noisy MS-COCO dataset to showcase its generalization ability. These results clearly indicate the great potential of data augmentation to handle noise annotations. {The source code is released at https://github.com/wscds/Mix-Paste.}
\end{abstract}


\section{Introduction}
\begin{figure}
  \includegraphics[width=0.9\linewidth]{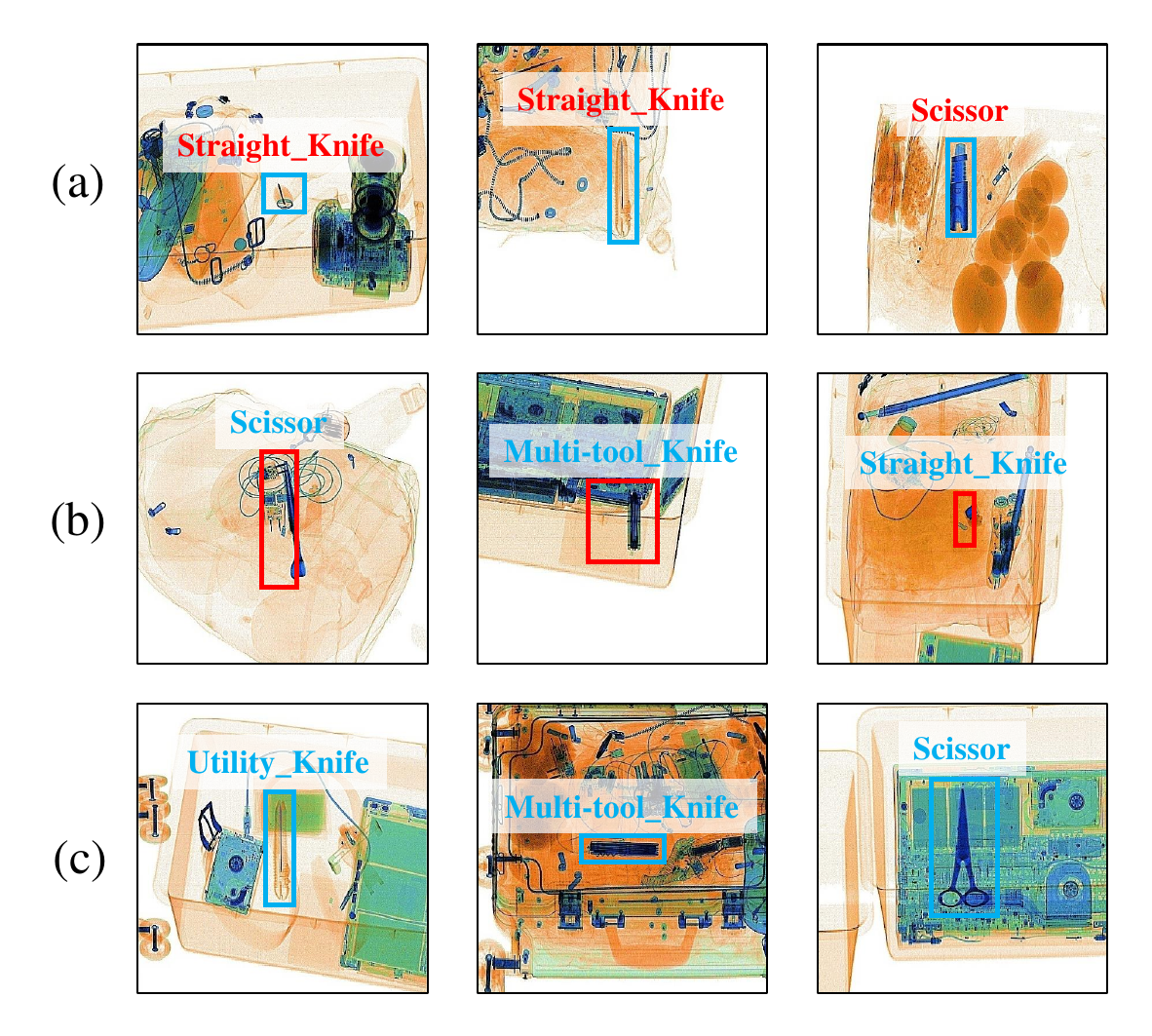}
  \caption{Examples in an X-ray dataset \cite{wei2020occluded}. (a) Examples with noisy category labels (the correct labels from left to right are folding knife, utility knife, and utility knife, respectively). (b) Examples with inaccurate bounding boxes. (c) Examples with correct annotations in X-ray images, where item overlapping is ubiquitous.}
  \label{fig:image}
\end{figure}
\IEEEPARstart{O}{ver} the past few years,  automatic X-ray prohibited item detection, which can assist security inspectors to quickly identify the locations and categories of prohibited items, has attracted much attention. A large number of prohibited item detection methods \cite{miao2019sixray, wei2020occluded, zhang2023pidray, tao2021towards,Griffin2019Unexpected,Ma2024Toward,yang2024dual} have been developed. 

Generally, existing X-ray prohibited item detection methods 
depend heavily on a large-scale dataset for model training. Unfortunately, obtaining correct annotations with clean category labels as well as accurate bounding boxes is labor-expensive and requires the expertise of professionals.
Notably, the ubiquitous item overlapping in X-ray images renders the annotation of an X-ray dataset a challenging task. 
{In many practical applications, 
machine-assisted annotations or crowd-sourcing 
are often employed to annotate large-scale data, reducing the expensive cost of high-quality human annotations. The machine-assisted process or the crowd-sourcing labeling 
process easily leads to noisy annotations.}
%The annotations
%can be noisy due to various reasons, including using crowd-sourcing or machine-generated annotations. Moreover, the ubiquitous item overlapping in X-ray images renders the annotation of an X-ray dataset a challenging task, increasing the probability of noisy annotations.}
{As a result, some existing X-ray datasets involve annotations with both \textit{category noise} (i.e., noisy category labels) and \textit{bounding box noise} (i.e., inaccurate ground-truth bounding boxes).} Fig. \ref{fig:image} gives some examples with noisy annotations in a public X-ray dataset \cite{wei2020occluded}. 
These noisy annotations greatly decrease the model performance.

To address the problem of learning with label noise,  existing methods \cite{li2020dividemix, han2018co, wang2019symmetric} often adopt a {label refinement} or loss correction paradigm. However, most of these methods work on the image classification task without considering the existence or the location of the objects/items. 
Unlike the image classification task,  the X-ray prohibited item detection task introduces additional challenges caused by inaccurate ground-truth bounding boxes. As a consequence, label noise learning methods do not work well on the prohibited item detection task (a specialized task of object detection). 
Recently, some works \cite{chadwick2019training, li2020towards, yang2020learning, liu2022robust, wang2022narrowing} have studied object detection under noisy annotations. 
But they are designed for {common} object detection and are not well-suited for prohibited item detection due to the presence of ubiquitous item overlapping in X-ray images.

To effectively train a robust prohibited item detector in noisy scenarios, we revisit the fundamental aspect of learning with noisy annotations (i.e., reducing the noise during training) and the inherent characteristics of X-ray images (i.e., the ubiquitous overlapping between items) from the perspective of data augmentation. {In particular, for an X-ray dataset contaminated with noisy annotations, a collection of item patches that {share} the same category label is more likely to contain one correct prohibited item than the individual patch in the
collection. Inspired by this observation, we mix such multiple item patches
to generate a mixed patch and paste it back into the original image for data augmentation. 
Thus, the generated image can involve the correct prohibited item with a high probability, reducing the negative influence of noisy annotations. 
In fact, the mixing process of these multiple item patches also effectively mimics item overlapping in X-ray images, enabling the detector to gain
a deeper understanding of X-ray images and improve the detection performance.} 

Although the mixed patch can effectively alleviate the noisy annotations, it may introduce additional noisy-labeled (caused by category noise) prohibited items during training. In such a case, the model tends to give predictions for all the possible prohibited items in the mixed patch before overfitting noisy labels.  When the conventional classification loss is used for model optimization, some accurate predictions of additional prohibited items may be mistakenly considered as false predictions (since the category label of additional prohibited items is noisy) and generate large losses. Such a way can be harmful to model training. Hence, we should remove the large losses corresponding to these potentially positive predictions during loss calculation.

Based on the above analysis, 
we propose a simple yet effective data augmentation method,
called label-aware mixed patch paste augmentation (\textbf{Mix-Paste}) to address the problem of training a robust X-ray prohibited item detector under noisy annotations.  
Specifically, for each item patch (corresponding to a ground-truth bounding box) in the training {image}, we first randomly choose several item patches (according to their ground-truth bounding boxes) with the same category label from different images. Then, we mix these patches and {replace the original patch in the image with the mixed patch, obtaining a new image.} 
By doing this, the probability of containing the correct prohibited item within the generated image is increased. {It is worth pointing out that our method randomly selects 
item patches with the same category label, where the category label of some patches can be contaminated with noise. In other words, it does not require 
the label of the selected item patches to be clean. In fact, such a selection can increase the probability of containing the correct prohibited item in the mixed patch.}
To effectively optimize the model on augmented data,  we design an item-based large-loss suppression (LLS) strategy,  suppressing the large losses corresponding to potentially positive predictions of additional items caused by the mixing operation in Mix-Paste.

\noindent
In summary, our contributions are given as follows:
\begin{itemize}
    \item{We propose a new data augmentation method 
    by mixing item patches with the same category label for X-ray prohibited item detection. 
    Our method can significantly reduce category noise and bounding box noise during training, obtaining a noise-robust prohibited item detector. 
{To the best of our knowledge, we are the first to address the problem of noisy annotations in X-ray prohibited item detection from a novel  perspective of data augmentation.}}
    \item{We design an effective loss suppression strategy for loss calculation. Such a strategy 
overcomes the limitations of the small-loss criterion \cite{zhang2021understanding, arpit2017closer}  
for label noise learning in our task. 
This greatly reduces the adverse influence of the large losses corresponding to potentially positive predictions of additional items} caused by the mixing process of item patches and enhances the detection performance.
\end{itemize}
Our extensive experiments on public X-ray datasets validate the effectiveness of our method under noisy annotations. We also perform experiments on the noisy MS-COCO dataset \cite{lin2014microsoft}, which exists a certain level of object overlapping.
Interestingly, our method shows performance improvements on the {common} object detection task. These results clearly indicate the advantage of data augmentation to address the problem of learning with noisy annotations. 

The remainder of this paper is organized as follows. We first review the related work in Section \ref{relatedworks}. Then, we elaborately introduce our proposed method in Section \ref{methodology}. Next, we conduct the experiments on the noisy X-ray datasets and the noisy MS-COCO dataset in Section \ref{experiments}. Finally, we draw the conclusion in Section \ref{conclusion}.

\section{Related Works}
\label{relatedworks}
{In this section, we briefly review several related works. We first introduce X-ray prohibited item detection methods in Section \ref{sec:detection}. 
{Then, we review data augmentation methods in Section \ref{sec:label-aug}.}
Finally, we review the methods of learning with label noise and learning with noisy annotations for object detection in Section \ref{sec:label-noise} and Section \ref{sec:noisyann},  respectively.}

\subsection{X-Ray Prohibited Item Detection}\label{sec:detection}
With the development of deep learning technology, 
automatic X-ray prohibited item detection has been widely applied in security inspection. {A variety of  methods\cite{wei2020occluded, tao2021towards,  zhang2023pidray,Ma2024Toward,zhao2022detecting,shao2022exploiting,9842976,akcay2022towards,10005308,velayudhan2022recent,10322651} have been developed to 
address the severe occlusion and item overlapping problems in X-ray images by introducing attention mechanisms or specifically-designed modules.} 
% \textcolor{red}{
%  Webb \textit{et al.} \cite{webb2021operationalizing} investigates two end-to-end CNN object detection architectures for prohibited item detection, examines the impact of lossy image compression on CNN-based detection.}
Wei \textit{et al.} \cite{wei2020occluded} propose an attention mechanism to enhance the edge and material information of prohibited items. 
Zhang \textit{et al.} \cite{zhang2023pidray} apply spatial- and channel-wise attention mechanisms to extract discriminative features and incorporate a dependency refinement module to explore long-range dependencies within the feature map. 
Tao \textit{et al.} \cite{tao2021towards} identify the object regions for prohibited item detection by removing the noisy information from neighboring regions and activating the boundary information. 
{Ma \textit{et al.} \cite{Ma2024Toward} leverage dual-view X-ray images as the input and exploit non-overlapping information of two images to enhance feature representations of prohibited items, effectively mitigating background overlapping.}
{Zhao \textit{et al.} \cite{zhao2022detecting} introduce a label-aware mechanism to tackle the item overlapping problem by establishing the associations between feature channels and labels. Based on this, they refine and adjust the features to enhance prediction results according to the assigned  pseudo labels.} 
{
Shao \textit{et al.} \cite{shao2022exploiting} propose a foreground and background separation (FBS) method, which can handle the severe overlapping problem in X-ray images by separating prohibited items from other irrelevant items. 
Velayudhan \textit{et al.} \cite{9842976} introduce a baggage threat detection framework based on broad learning. This framework leverages low-rank features to identify and localize concealed and cluttered baggage threats.}

Due to the ubiquitous item overlapping in X-ray images, annotating an X-ray dataset becomes a challenging task. 
{As a result, some X-ray datasets involve noisy annotations with both {category noise} and {bounding box noise}. Hence, it is crucial to develop a noise-robust prohibited item detector.}

{
\subsection{Data Augmentation}\label{sec:label-aug}}

{Data augmentation aims to improve the 
generalization capability of models by artificially increasing the diversity of the training data. Cutout \cite{devries2017improved} randomly applies masking to square patches in the image, effectively enforcing the model to learn from incomplete information. Mix-Up \cite{zhang2017mixup} generates new training examples by linearly interpolating between two images and their corresponding labels. This encourages the model to generalize beyond the training data and reduce sensitivity to adversarial examples. 
AlignMix\cite{venkataramanan2022alignmixup} improves representation learning by geometrically aligning and interpolating features from multiple images. Such a way enhances the model's generalization and robustness.
Mosaic \cite{bochkovskiy2020yolov4} combines four images into one, enabling the model to observe multiple contexts in a single training step. This method increases the diversity of the dataset and exposes the model to more complex, multi-object scenes, enhancing robustness to variations in object scale and occlusions. Channel augmentation \cite{10319076} explores the relationship between visible and infrared images to obtain modality-invariant features.}

{
Recently, some methods leverage  CLIP\cite{radford2021learning} or the diffusion model\cite{zhang2023adding} to generate new data by using prompt words. Fang \emph{et al.} \cite{fang2024data} propose a data augmentation pipeline based on controllable diffusion models and CLIP for object detection. %It generates visual priors, constructs prompts, uses a controllable diffusion model to generate synthetic data, and performs post-filtering with category-calibrated CLIP scores. 
Gannamaneni \emph{et al.} \cite{gannamaneni2024exploiting} generate safety critical scenes by inpainting with diffusion models conditioned on text and pose, offering fine-grained control over pedestrian attributes.}

{Some of the above methods perform well in clean X-ray datasets \cite{webb2021operationalizing}. However, when applied to the X-ray datasets involving noisy annotations, the generated X-ray images contain significant disturbances,  potentially degrading model performance. Different from the above data augmentation methods, we design a data augmentation method to effectively alleviate noisy annotations in the X-ray dataset and improve the training performance of the model in the noisy dataset.}

\subsection{Learning with Label Noise}\label{sec:label-noise}
A number of methods \cite{li2020dividemix, han2018co, wang2019symmetric, zhang2018generalized, tan2021co, zhang2021understanding} have been developed for learning with label noise. 
Some methods \cite{goldberger2016training, patrini2017making, xia2020part, li2021provably} address the label noise problem by employing a noise transition matrix to refine predictions.
Goldberger \textit{et al.} \cite{goldberger2016training} adopt both an s-model and a c-model to effectively obtain a noise transition matrix. 
Patrini \textit{et al.} \cite{patrini2017making} explicitly model the noise transition matrix to correct the loss. 
Several works \cite{ghosh2017robust, zhang2018generalized, wang2019symmetric, ma2020normalized} reveal that a loss function involving symmetric properties exhibits enhanced robustness against label noise. 
However, these methods may only be capable of handling certain noisy rates. 
Recent methods focus on new learning paradigms \cite{jiang2018mentornet, han2018co, yu2019does, wei2020combating, kang2019decoupling,purifyNet,9640485,9725265}. For example, 
MentorNet \cite{jiang2018mentornet} leverages a teacher-student framework to learn a robust student model by exploiting the knowledge of a teacher model.
Co-teaching \cite{han2018co} trains the two models simultaneously, where each model selects the samples with the small-loss criterion to update the other model.
Co-teaching+ \cite{yu2019does} improves the performance of Co-teaching by training on disagreement data.
JoCoR \cite{wei2020combating} allows the two models to reach an agreement by minimizing the distance loss predicted by the two models.
{PurifyNet \cite{purifyNet} introduces a hard-aware instance re-weighting strategy to focus on hard samples in the noisy dataset.} {
Ye \textit{et al.} \cite{9640485} propose an online label co-refinement framework, which progressively refines noisy labels during model optimization.}

The above methods mainly handle label noise and focus on the image classification task. In this paper, our method addresses noisy annotations involving both category noise and bounding box noise for training a robust prohibited item detector.  {Moreover, unlike the above methods that select clean data by designing different strategies, our method addresses noisy annotations from the perspective of data augmentation. In this way, our method does not require  
 estimating/predefining the noise rates or selecting a clean subset as did in conventional label noise learning methods.}

 \subsection{Learning with Noisy Annotations for Object Detection}\label{sec:noisyann}
Recently, some methods \cite{chadwick2019training, li2020towards, yang2020learning, liu2022robust} have been developed to address robust training on noisy annotations for object detection.
Chadwick  \emph{et al.} \cite{chadwick2019training} extend the Co-teaching strategy to the field of object detection.
Li \emph{et al.} \cite{li2020towards} decouple bounding box noise from category noise and then leverage the predicted output for category noise correction and bounding box refinement.
Yang \emph{et al.} \cite{yang2020learning} reduce the influence of noisy labels by employing diverse loss functions.
Liu \emph{et al.} \cite{liu2022robust} treat each object as a bag of instances and select accurate instances from the object bags for training.
{Wang \emph{et al.} \cite{wang2022narrowing} develop a novel Bayesian filter-based prediction ensemble method to address noisy bounding box annotations within a teacher-student learning framework.}

The aforementioned methods are designed for {common} object detection. In contrast, 
our method mixes multiple item patches to mimic the characteristics of X-ray images (i.e., the ubiquitous overlapping between  items)
for prohibited item detection. Surprisingly, experiments show that our method is also beneficial in improving the performance of common object detection (which exists a certain level of object overlapping). 

\section{Methdology}
\label{methodology}
\begin{figure*}
    \centering
    \includegraphics[height=7.5cm, width=16cm]{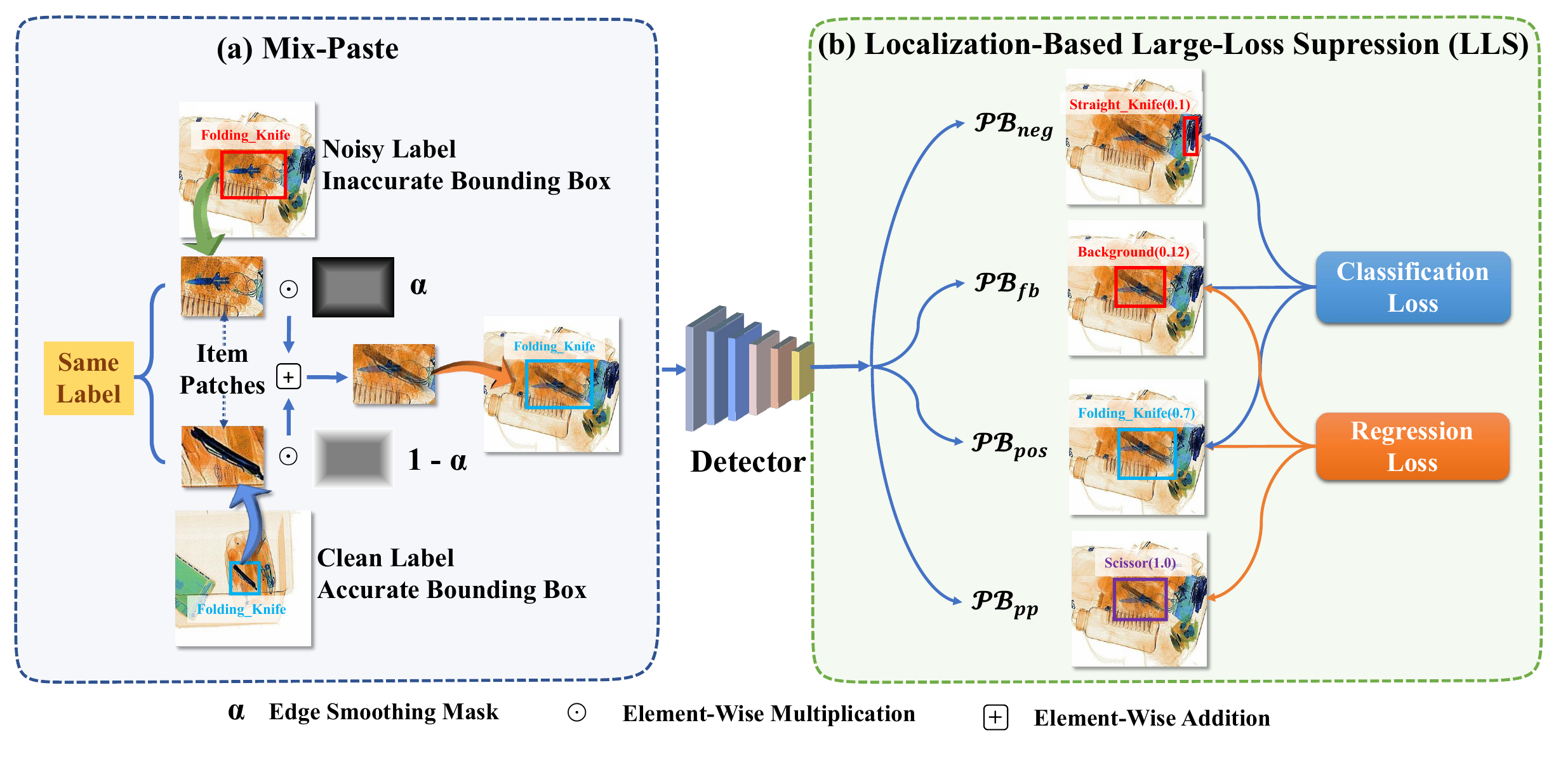}
    \caption{ Overview of our proposed method for training a robust prohibited item detector under noisy annotations. (a) illustrates Mix-Paste which
mixes multiple item patches with the same category label (the correct label in the upper item patch is the scissor) for data augmentation. {(b) illustrates
the LLS strategy which suppresses the large losses corresponding to potentially positive predictions of additional items during loss calculation. $\mathcal{PB}_{neg}$: the predicted bounding
boxes whose IoUs between them and the ground-truth bounding boxes are less than a threshold; $\mathcal{PB}_{fb}$: the predicted bounding boxes whose IoUs are greater than a threshold and the predicted label is the background; 
$\mathcal{PB}_{pos}$:  the predicted bounding boxes whose IoUs are greater than a threshold and the predicted label is the same as the ground-truth category label; $\mathcal{PB}_{pp}$: the predicted bounding boxes whose IoUs are greater than a threshold and the predicted label (not the background) is different from the ground-truth category label.}}
    \label{fig:mixpaste}
\end{figure*}

In this section, we first give the problem formulation in Section \ref{problem}. Then, we provide an overview of our method in Section \ref{sec:overview}. Next, we describe our Mix-Paste method in detail in Section \ref{sec:mix-paste}. Finally, we introduce an LLS strategy, which can be effectively combined with Mix-Paste to alleviate the influence of noisy annotations during model training, in Section \ref{sec:LLS}.

\subsection{Problem Formulation}
\label{problem}
{Some X-ray datasets involve noisy annotations} due to the difficulty of obtaining high-quality human annotations in X-ray images, where item overlapping is prevalent. In this paper, we address the problem of training a robust prohibited item detector on the noisy X-ray dataset, where the noise contains a mixture of category noise
and bounding box noise. In addition, we do not assume that a subset with clean annotations is available. 

Given a noisy dataset $\mathcal{D}=\{(\textbf{x}_i,\widetilde{\textbf{y}}_i)\}_{i=1}^N$, where $\textbf{x}_i$ is the $i$-the training image and $\widetilde{\textbf{y}}_i=\{\widetilde{{c}}_j, \widetilde{\textbf{b}}_j\}_{j=1}^{J_i}$ denotes 
the annotation of $\textbf{x}_i$. Here, $\widetilde{{c}}_j$ denotes the label of the $j$-th prohibited item,  $\widetilde{\textbf{b}}_j=(\widetilde{x}, \widetilde{y}, \widetilde{w}, \widetilde{h})$ represents the ground-truth bounding box coordinates ($\widetilde{x}$ and  $\widetilde{y}$ represent the coordinates of the top-left corner, and $\widetilde{w}$ and $\widetilde{h}$ represent the width and height, respectively)  of the $j$-th prohibited item, and $J_i$ is the number of prohibited items for $\textbf{x}_i$.
Unlike the image classification task, the prohibited item detection task often involves two types of noise: {category noise} and {bounding box noise}. In this way, the dataset contains class-corrupted instances where the category labels are noisy, and position-corrupted instances where the ground-truth bounding boxes are inaccurate. Hence, we aim to train a noise-robust model based on $\mathcal{D}$ and evaluate its performance on the test set. 

\subsection{Overview}\label{sec:overview}
In this paper, we develop a simple yet effective data augmentation method called Mix-Paste for training on the noisy X-ray dataset. Mix-Paste is a plug-and-play data augmentation method that can be directly applied to the training of different prohibited item detectors.  The overview of our method is shown in Fig. \ref{fig:mixpaste}.

Specifically, for each item patch corresponding to a ground-truth bounding box in the training image, we first randomly select several item patches (specified by the {ground-truth bounding boxes} with the same category label) from different images. Then, we resize these item patches to the same size and mix them. Finally, we can paste the mixed patch back into the original item location in the image. 
In this way, the mixture of item patches can generate a new training image with reduced noise interference. {Note that instead of applying Mix-Paste to all the images, we apply our proposed Mix-Paste to the randomly selected subset of the training set  (with a probability),  maintaining consistency between the training and test samples. In this way, some X-ray images are generated by Mix-Paste while the other images are unchanged.}

To obtain a robust detector on the augmented data, we further design an item-based large-loss suppression (LLS) strategy, which suppresses the large losses corresponding to potentially positive predictions of additional items  during loss calculation. 
Technically, we first select the predicted bounding boxes, for which the Intersection over Unions (IoUs) between them and the ground-truth bounding boxes 
are larger than a threshold. Then, we identify those predicted bounding boxes whose  predicted label is different from the ground-truth category label (except for predicted bounding boxes whose  predicted label is the background). Finally, the classification losses
corresponding to these identified bounding boxes are suppressed and not counted for loss calculation. By doing this, we effectively reduce the negative influence caused by the patch mixing operations (which may involve several different prohibited items due to category noise) for model optimization. 

\subsection{Mix-Paste}\label{sec:mix-paste}
%\textcolor{red}{For each item patch which will apply the Mix-Paste augmentation, 
%and corresponding to the ground-truth bounding box in the image $\textbf{x}$, 
 {We generate a new patch by mixing multiple item patches that share the same category label.}
%\st{(except for the original image)}.
To mix these item patches, we crop the item 
patches from the selected images according to the {ground-truth bounding boxes} and resize them to the same size.
Finally,  the mixed patch is used to replace the original patch. Note that we only randomly select patches with the same category label from the whole dataset without assuming that the labels of the selected patches are clean. 

Mathematically, the process of generating the mixed patch { $\hat{\mathbf{B}}$ }is formulated as
{
\begin{equation}
    \hat{\mathbf{B}} = \bm{\alpha} \odot \mathbf{B}_a + \sum_{n = 2}^{K}  \frac{1-\bm{\alpha}}{K - 1} \odot \text{resize}(\mathbf{B}_n),
\end{equation}
}
where $K$ is the total number of patches for mixing {(including the original item patch)}; 
{$\mathbf{B}_a$} is the original item patch in $\textbf{x}$; {$\mathbf{B}_n$} is the $n$-th item patch randomly selected from the whole dataset;  `$\odot$' is the element-wise multiplication; $\rm{resize(\cdot)}$ denotes the function that resizes the item patch to the same size as {$\mathbf{B}_a$};
{$\bm{\alpha}$} is an edge smoothing mask to make the mixed patch more natural.

The edge smoothing mask is defined as 
%The gradient edge mask is generated by the following formula:
{
\begin{equation}
    \bm{\alpha}(i,j) =\left\{
      \begin{aligned}
      &1 - (1 - \lambda) (\bm{d}_{i,j} /(\beta \cdot w)) , & \bm{d}_{i,j} \leq \beta \cdot w, \\
      &\lambda, & \bm{d}_{i,j} > \beta \cdot w, \\
      \end{aligned}
      \right.
\end{equation}
}
where $\bm{d}_{i,j}$ is the distance between the pixel (with the spatial location of ($i$,$j$) in the patch) to the nearest boundary of the patch; $\beta$ denotes a threshold to control the smoothing area (we empirically set $\beta$ to 10\%); {$w$ is the width of the bounding box;} $\lambda \in [0, 1]$ is a random number generated from a Beta distribution. Although the edge smoothing mask can allow for the natural appearance of the mixed patch, we also observe that simply merging the patches with linear combinations can also achieve similar performance. 

{Note that some methods apply threat image projection for image fusion. However, it is difficult to apply threat image projection in our method due to the following several reasons.
First, some threat image projection methods \cite{bhowmik2019good,duan2023rwsc} require X-ray images with plain backgrounds to segment  prohibited items and superimpose isolated
prohibited items onto normal images. As most X-ray datasets do not have X-ray images with plain backgrounds, it is not trivial to obtain isolated prohibited items.
Second, traditional threat image projection methods \cite{mery2017logarithmic,rogers2016threat} only work on the fusion of gray images. However, most current X-ray datasets are color images (notice that the color information of each prohibited item plays an important role in detection due to the penetration characteristics of X-rays). Therefore, these methods cannot be directly used in our method. Although recent methods \cite{duan2023rwsc} extend threat image projection to color X-ray images by superimposing pixel-level isolated prohibited items onto normal images, the pixel-level annotations are not available in our X-ray datasets. 
Third, most threat image projection methods 
compute the image intensity based on the X-ray energy, object material, and object thickness. 
In this way, some parameters related to the X-ray scanners (such as the X-ray energy) are required as a prerequisite. But these parameters are not provided in existing public X-ray datasets.}

Subsequently, the mixed patch is pasted back into the original image, which can be formulated as

\begin{equation}
  \textbf{x}{[\widetilde{x}: \widetilde{x} + \widetilde{w}, \widetilde{y}:\widetilde{y} + \widetilde{h}]} = \hat{\mathbf{B}},
\end{equation}

{where $\textbf{x}$ denotes the original image corresponding to the item patch;}
$[\widetilde{x}:\widetilde{x} + \widetilde{w}, \widetilde{y}:\widetilde{y} + \widetilde{h}]$ denote the bounding box region of {the original item patch ${\mathbf{B}_a}$}.

\noindent \textbf{Why does Mix-Paste work?} 
We analyze the reasons why our Mix-Paste can work on the training of X-ray prohibited item detection under noisy annotations. First, Mix-Paste can reduce category noise and bounding box noise explicitly. 
For an annotated bounding box with the prohibited item label  $\widetilde{{c}}_j$ in the dataset involving the category noise rate of $P_c$, the probability of the prohibited item within the bounding box region is estimated as $1-P_c$. 
When $K$ item patches with the same category label $\widetilde{{c}}_j$ are mixed, the probability of the existence of the item with the label $\widetilde{{c}}_j$ (which is computed as { $1-P_{c}^{K}$}) is increased. %, where $K$ is the number of mixed patches. 
Analogously, suppose that the bounding box noise rate is $P_b$, {the probability of the $K$ mixed patch that can accurately bound a correct prohibited item (which is computed as $1-P_{b}^{K}$) is also increased.}
Second, Mix-Paste can effectively mimic item overlapping in X-ray images, 
thereby enabling the detector to enhance its awareness of overlapping.
Third, Mix-Paste can generate more diverse training samples, thereby enhancing the generalization ability of the model. 

\noindent \textbf{Can the mixing operation perfectly mimic item overlapping in X-ray images?}
{Some existing data augmentation methods (such as Mix-Up \cite{zhang2017mixup}/CutMix \cite{yun2019cutmix}) fail to generate data perfectly as the original dataset. However, these data augmentation
methods can significantly enhance the model performance in various tasks by substantially increasing the diversity of the dataset. In the same spirit, although the mixing operation in Mix-Paste cannot perfectly mimic item overlapping in X-ray images,  it still can encourage the model to learn some characteristics of X-ray images under item overlapping conditions. More importantly, our Mix-Paste is shown to be effective in alleviating the influence of noisy annotations during model training.}

\noindent {\textbf{Can Mix-Paste be applied to the segmentation task?}
Unfortunately, our method is difficult to 
be applied to the segmentation task. The core idea behind our method is to increase the probability of the target prohibited item appearing within a bounding box by mixing different item patches. It is 
straightforward to adjust the different sizes of bounding boxes for patch mixing since these boxes are rectangular. However, 
it is not easy to align the object with different shapes at the pixel level for the  segmentation task.  As a result, our method is more suitable for object detection than segmentation.}

\noindent \textbf{{Comparison 
against traditional data augmentation methods.}}
%against Mix-Up/CutMix.} 
{Both our method and traditional data augmentation methods combine the training samples to generate new samples. However, there are some intrinsic differences (in terms of the motivations and methodological details) between our method and traditional methods. 
First, %the motivations of the two methods are different. 
Mix-Up encourages
the model to behave linearly, reflecting a good inductive
bias \cite{zhang2017mixup}. In contrast, Mix-Paste aims to synthesize more training samples with less annotation noise.
Second, Mix-Up, which combines the two images at the image level, is developed for image classification. On the contrary, Mix-Paste, 
which mixes item patches with the same category label at the patch level, is designed for prohibited item detection. Our experiments further validate that Mix-Paste significantly outperforms Mix-Up for prohibited item detection under noisy annotations. 
CutMix \cite{yun2019cutmix} randomly replaces a patch in the image with another patch from another image.
SaliencyMix\cite{uddin2020saliencymix} and Attentive CutMix\cite{walawalkar2020attentive} enhance CutMix by pasting the most salient region %, identified from the saliency map, 
onto the corresponding location in the target image. 
Unlike the above methods, Mix-Paste mixes multiple item patches with the same category label.} %Note that traditional data augmentation methods are mainly designed for image classification, while our MixPaste is specifically designed for addressing prohibited item detection with noisy annotations.} 

\subsection{Item-Based Large-Loss Suppression (LLS) Strategy}\label{sec:LLS}
{After the mixing operation, the probability of the mixed patch containing the correct target prohibited item is increased (the detailed analysis about why
our Mix-Paste can work on the training of X-ray prohibited item detection under noisy annotations is given in Section III-C).
However, the mixing operation is likely to introduce additional noisy-labeled prohibited items (caused by category noise in some selected patches)  during training. 
In fact, the probability that the selected patches consist of all correct prohibited items is only $(1-P_c)^K$, where $P_c$ denotes the category noise rate and $K$ is the number of patches. Consequently, the mixing operation may introduce additional noisy-labeled  prohibited items during training. }
%delete After applying Mix-Paste, the probability that the generated image contains the correct prohibited item increases. 
% But the generated image may contain additional noisy-labeled prohibited items from other categories due to category noise. 
In such a case, the model tends to predict these items for the newly generated image during training. However, these predicted bounding boxes  
will be mistakenly considered as false predictions since their corresponding correct labels are not available. 
Hence, these potentially positive predictions give large losses in the conventional classification loss calculation,  resulting in a negative influence on model training.

To alleviate this problem, we propose an item-based large-loss suppression (LLS) strategy. 
{As illustrated in Fig.~\ref{fig:mixpaste}}, we categorize the  prediction results into four parts, including (1) { (1) $\mathcal{PB}_{neg}$: the predicted bounding
boxes whose IoUs between them and the ground-truth bounding boxes are less than a threshold (e.g., there is no matching between the ground-truth bounding box and the predicted bounding box in Fig.~2(b));  (2) $\mathcal{PB}_{fb}$: the predicted bounding boxes whose IoUs are greater than a threshold and the predicted label is the background (e.g., the predicted bounding box position is correct but the category label is predicted to the background in Fig.~2(b));
(3) $\mathcal{PB}_{pos}$:  the predicted bounding boxes whose IoUs are greater than a threshold and the predicted label is the same as the ground-truth category label (e.g., both the predicted bounding box position and category label are correct in Fig.~2(b)); (4) $\mathcal{PB}_{pp}$: the predicted bounding
boxes whose IoUs are greater than a threshold and the predicted label (not the background) is different from the ground-truth category label (e.g., the predicted bounding box position is correct but the predicted category label is incorrect in Fig.~2(b)).}

When calculating the classification loss, we suppress $\mathcal{PB}_{pp}$ from the prediction results and focus on the remaining three parts. 
Hence, the final loss is calculated as
\begin{equation}
    \mathcal{L} = \mathcal{L}_{bbox} + \mathcal{L}_{cls_{neg}} + \mathcal{L}_{cls_{pos}} + \mathcal{L}_{cls_{fb}},
\end{equation}
where $\mathcal{L}_{bbox}$ denotes the bounding box regression loss; $\mathcal{L}_{cls_{neg}}$, $\mathcal{L}_{cls_{fb}}$, and $\mathcal{L}_{cls_{pos}}$ denote the classification losses for $\mathcal{PB}_{neg}$, $\mathcal{PB}_{fb}$, and  $\mathcal{PB}_{pos}$, respectively.

For learning with label noise on image classification, the popular small-loss criterion \cite{arpit2017closer, zhang2021understanding} treats samples with small losses as clean samples and considers samples with large losses as noisy samples.
However, for prohibited item detection, the loss calculation contains both foreground and background predictions, where the background predictions account for the majority of the total loss. 
As a result, the small-loss criterion mainly focuses on background predictions and may ignore foreground predictions in this task. In contrast, our LLS strategy is highly effective in handling category noise by removing only the potentially positive predictions. 

\noindent \textbf{Why does the LLS strategy work?} In the LLS strategy, we ignore
the predicted bounding boxes whose predicted labels are
different from the ground-truth category labels (except for
those whose predicted label is the background) when calculating the classification loss.
In noisy scenarios, the mixed patches may contain multiple prohibited items because of category noise. 
Consequently, when these newly generated images are used for training, the model tends to give correct predictions for additional items. 
However, since these items are not associated with correct labels, they are considered as false predictions and consequently give {large losses during model training. This will lead to incorrect model optimization, reducing the overall performance of the model.}
To mitigate this and enhance the model performance, we suppress the large losses corresponding to potentially positive predictions of additional items for loss calculation.
Note that we still compute the loss for those predicted bounding boxes whose predicted labels are the background since the predicted bounding box region contains a prohibited item.

\section{experiments}
\label{experiments}
{In this section, we first introduce the datasets and evaluation metrics in Section \ref{sec:datasets}. {Then, we present the noise rate estimation and implementation details of our method in Section \ref{sec:estimation} and Section \ref{sec:detail}, respectively.} Next, we compare our method with state-of-the-art methods on the noisy X-ray datasets in Section \ref{sec:exp-xray} and the noisy MS-COCO dataset in Section \ref{sec:exp-coco}. After that, we conduct ablation studies in Section \ref{sec:ablation}. Finally, we give some visualization results in Section \ref{sec:vis}.}

\subsection{Datasets}\label{sec:datasets}

\begin{table*}[!t] \scriptsize
  
  \caption{Comparison Results (\%) on the OPIXray Dataset. $P_c$ and $P_b$ Denote the Category Noise Rate and Bounding Box Noise Rate, Respectively. }
  \centering
  \setlength{\tabcolsep}{4mm}{
  \begin{tabular}{lllllll}
    \toprule
    \multicolumn{1}{c}{\multirow{2}{*}{Method}} &  \multicolumn{2}{c}{$P_c=20\% \quad P_b=20\%$} &\multicolumn{2}{c}{$P_c=40\% \quad P_b=40\%$} & \multicolumn{2}{c}{$P_c=60\% \quad P_b=60\%$} \\
    \cmidrule(lr){2-3} \cmidrule(lr){4-5} \cmidrule(lr){6-7} 
      & \multicolumn{1}{c}{mAP@.5} & \multicolumn{1}{c}{mAP@[.5, .95]} & \multicolumn{1}{c}{mAP@.5} & \multicolumn{1}{c}{mAP@[.5, .95]} & \multicolumn{1}{c}{mAP@.5} & \multicolumn{1}{c}{mAP@[.5, .95]} \\
     \midrule
     FRCNN (PAMI, '17) \cite{ren2015faster}  & 81.1 (+0.0) & 31.5 (+0.0) & 70.0 (+0.0) & 25.7 (+0.0) & 56.7 (+0.0) & 18.4 (+0.0) \\
     LIM (ICCV, '21) \cite{tao2021towards}  & 83.7 \textcolor[rgb]{0.1961, 0.8039, 0.1961}{(+2.6)} & 34.6 \textcolor[rgb]{0.1961, 0.8039, 0.1961}{(+3.1)} & 80.2 \textcolor[rgb]{0.1961, 0.8039, 0.1961}{(+10.2)} & 31.3 \textcolor[rgb]{0.1961, 0.8039, 0.1961}{(+5.6)} & 72.7 \textcolor[rgb]{0.1961, 0.8039, 0.1961}{(+16.0)} & 26.4 \textcolor[rgb]{0.1961, 0.8039, 0.1961}{(+8.0)}\\
     SDANet (IJCV, '23) \cite{zhang2023pidray}  & 83.9 \textcolor[rgb]{0.1961, 0.8039, 0.1961}{(+2.8)} & 32.6 \textcolor[rgb]{0.1961, 0.8039, 0.1961}{(+1.1)} & 71.2 \textcolor[rgb]{0.1961, 0.8039, 0.1961}{(+1.2)} & 25.1 \textcolor{red}{(-0.6)} & 52.4 \textcolor{red}{(-4.3)} & 17.1 \textcolor{red}{(-1.3)}\\
     {GADet} (SENS J., '24) \cite{10322651} & 81.2 \textcolor[rgb]{0.1961, 0.8039, 0.1961}{(+0.1)} & 34.5 \textcolor[rgb]{0.1961, 0.8039, 0.1961}{(+3.0)} & 77.5 \textcolor[rgb]{0.1961, 0.8039, 0.1961}{(+7.5)} & 32.5 \textcolor[rgb]{0.1961, 0.8039, 0.1961}{(+6.8)} & 69.7 \textcolor[rgb]{0.1961, 0.8039, 0.1961}{(+13.0)} & 27.9 \textcolor[rgb]{0.1961, 0.8039, 0.1961}{(+9.5)}\\
     SCE (ICCV, '19) \cite{wang2019symmetric} & 81.8 \textcolor[rgb]{0.1961, 0.8039, 0.1961}{(+0.7)} & 32.5 \textcolor[rgb]{0.1961, 0.8039, 0.1961}{(+1.0)} & 72.1 \textcolor[rgb]{0.1961, 0.8039, 0.1961}{(+2.1)} & 25.2 \textcolor{red}{(-0.5)} & 48.6 \textcolor{red}{(-8.1)} & 16.1 \textcolor{red}{(-2.3)}\\
     LNCIS (ECCV, '20) \cite{yang2020learning} & 84.3 \textcolor[rgb]{0.1961, 0.8039, 0.1961}{(+3.2)} & 34.2 \textcolor[rgb]{0.1961, 0.8039, 0.1961}{(+2.7)} & 80.4 \textcolor[rgb]{0.1961, 0.8039, 0.1961}{(+10.4)} & 29.6 \textcolor[rgb]{0.1961, 0.8039, 0.1961}{(+3.9)} & 65.9 \textcolor[rgb]{0.1961, 0.8039, 0.1961}{(+9.2)} & 23.4 \textcolor[rgb]{0.1961, 0.8039, 0.1961}{(+5.0)} \\
     OA-MIL (ECCV, '22) \cite{liu2022robust}  & 82.2 \textcolor[rgb]{0.1961, 0.8039, 0.1961}{(+1.1)} & 31.3 \textcolor{red}{(-0.2)} & 70.2 \textcolor[rgb]{0.1961, 0.8039, 0.1961}{(+0.2)} & 27.5 \textcolor[rgb]{0.1961, 0.8039, 0.1961}{(+1.8)} & 56.4 \textcolor{red}{(-0.3)} & 21.6 \textcolor[rgb]{0.1961, 0.8039, 0.1961}{(+3.2)} \\
     Ours & \textbf{87.0 \textcolor[rgb]{0.1961, 0.8039, 0.1961}{(+5.9)}} & \textbf{38.3 \textcolor[rgb]{0.1961, 0.8039, 0.1961}{(+6.8)}} & \textbf{86.7 \textcolor[rgb]{0.1961, 0.8039, 0.1961}{(+16.7)}} & \textbf{37.0 \textcolor[rgb]{0.1961, 0.8039, 0.1961}{(+11.3)}} & \textbf{81.8 \textcolor[rgb]{0.1961, 0.8039, 0.1961}{(+25.1)}} & \textbf{33.7 \textcolor[rgb]{0.1961, 0.8039, 0.1961}{(+15.3)}} \\
    \bottomrule
  \end{tabular}}

  \label{tab:opi}
\end{table*}

\begin{table*}[!t]\scriptsize
  
  \caption{{Comparison Results (\%) on the PIDray Dataset. $P_c$ and $P_b$ Denote the Category Noise Rate and Bounding Box Noise Rate, Respectively.  We report mAP@[.5, .95] as the evaluation metric.}}
  \centering
  \setlength{\tabcolsep}{2.35mm}{
  \begin{tabular}{lllllllll}
    \toprule
    % \multicolumn{1}{c}{\multirow{2}{*}{Method}} &  \multicolumn{2}{c}{Original} & \multicolumn{2}{c}{$P_c=20\% \quad P_b=20\%$} &\multicolumn{2}{c}{$P_c=40\% \quad P_b=40\%$} & \multicolumn{2}{c}{$P_c=60\% \quad P_b=60\%$} \\
    
    \multicolumn{1}{c}{\multirow{2}{*}{Method}}  &\multicolumn{4}{c}{$P_c=40\% \quad P_b=40\%$} & \multicolumn{4}{c}{$P_c=60\% \quad P_b=60\%$} \\
    \cmidrule(lr){2-5} \cmidrule(lr){6-9}
     & \multicolumn{1}{c}{easy} & \multicolumn{1}{c}{hard} & \multicolumn{1}{c}{hidden} & \multicolumn{1}{c}{Avg} & \multicolumn{1}{c}{easy} & \multicolumn{1}{c}{hard} & \multicolumn{1}{c}{hidden} & \multicolumn{1}{c}{Avg}  \\
     \midrule
     FRCNN (PAMI, '17) \cite{ren2015faster} & 42.5 (+0.0) & 40.1 (+0.0) & 19.9 (+0.0) & 34.2 (+0.0) & 29.4 (+0.0) & 27.4 (+0.0) & 13.3 (+0.0) & 23.4 (+0.0)  \\
     LIM (ICCV, '21) \cite{tao2021towards} & 53.6 \textcolor[rgb]{0.1961, 0.8039, 0.1961}{(+11.1)} & 49.2 \textcolor[rgb]{0.1961, 0.8039, 0.1961}{(+9.1)} & 27.6 \textcolor[rgb]{0.1961, 0.8039, 0.1961}{(+7.7)} & 43.5 \textcolor[rgb]{0.1961, 0.8039, 0.1961}{(+9.3)} & 43.8 \textcolor[rgb]{0.1961, 0.8039, 0.1961}{(+14.4)} & 40.5 \textcolor[rgb]{0.1961, 0.8039, 0.1961}{(+13.1)} & 19.9 \textcolor[rgb]{0.1961, 0.8039, 0.1961}{(+6.6)} & 34.7 \textcolor[rgb]{0.1961, 0.8039, 0.1961}{(+11.3)}  \\
     SDANet (IJCV, '23) \cite{zhang2023pidray} & 40.8 \textcolor{red}{(-1.7)}  & 37.8 \textcolor{red}{(-2.3)} & 19.3 \textcolor{red}{(-0.6)} & 32.6 \textcolor{red}{(-1.6)} & 26.0 \textcolor{red}{(-3.4)} & 25.0 \textcolor{red}{(-2.4)} & 11.1 \textcolor{red}{(-2.2)} & 20.7 \textcolor{red}{(-2.7)}  \\
     {GADet} (SENS J., '24) \cite{10322651} & 47.9 \textcolor[rgb]{0.1961, 0.8039, 0.1961}{(+5.4)}  & 42.9 \textcolor[rgb]{0.1961, 0.8039, 0.1961}{(+2.8)} & 18.9 \textcolor{red}{(-1.0)} & 36.6 \textcolor[rgb]{0.1961, 0.8039, 0.1961}{(+2.4)} & 41.3 \textcolor[rgb]{0.1961, 0.8039, 0.1961}{(+11.9)} & 34.5 \textcolor[rgb]{0.1961, 0.8039, 0.1961}{(+7.1)} & 16.7 \textcolor[rgb]{0.1961, 0.8039, 0.1961}{(+3.4)} & 30.8 \textcolor[rgb]{0.1961, 0.8039, 0.1961}{(+7.4)}  \\
     SCE (ICCV, '19) \cite{wang2019symmetric} & 40.3 \textcolor{red}{(-2.2)} & 37.9 \textcolor{red}{(-2.2)} & 21.5 \textcolor[rgb]{0.1961, 0.8039, 0.1961}{(1.6)} & 33.2 \textcolor{red}{(-1.0)} & 24.7 \textcolor{red}{(-4.7)} & 23.3 \textcolor{red}{(-4.1)} & 11.1\textcolor{red}{(-2.2)} & 19.7\textcolor{red}{(-3.7)}  \\
     LNCIS (ECCV, '20) \cite{yang2020learning} & 48.2 \textcolor[rgb]{0.1961, 0.8039, 0.1961}{(+5.7)} & 45.2 \textcolor[rgb]{0.1961, 0.8039, 0.1961}{(+5.1)} & 25.9 \textcolor[rgb]{0.1961, 0.8039, 0.1961}{(+6.0)} & 39.8 \textcolor[rgb]{0.1961, 0.8039, 0.1961}{(+5.6)} & 33.4 \textcolor[rgb]{0.1961, 0.8039, 0.1961}{(+4.0)} & 30.9 \textcolor[rgb]{0.1961, 0.8039, 0.1961}{(+3.5)} & 14.5 \textcolor[rgb]{0.1961, 0.8039, 0.1961}{(+1.2)} & 26.3 \textcolor[rgb]{0.1961, 0.8039, 0.1961}{(+2.9)}  \\
     OA-MIL (ECCV, '22) \cite{liu2022robust} & 44.6 \textcolor[rgb]{0.1961, 0.8039, 0.1961}{(+2.1)} & 39.6 \textcolor{red}{(-0.5)} & 23.8 \textcolor[rgb]{0.1961, 0.8039, 0.1961}{(+3.9)} & 36.0 \textcolor[rgb]{0.1961, 0.8039, 0.1961}{(+1.8)} & 30.5 \textcolor[rgb]{0.1961, 0.8039, 0.1961}{(+1.1)} & 25.0 \textcolor{red}{(-2.4)} & 10.6 \textcolor{red}{(-2.7)} & 22.0 \textcolor{red}{(-1.4)}  \\
     Ours & \textbf{57.4 \textcolor[rgb]{0.1961, 0.8039, 0.1961}{(+14.9)}} & \textbf{53.2 \textcolor[rgb]{0.1961, 0.8039, 0.1961}{(+13.1)}} &  \textbf{31.9 \textcolor[rgb]{0.1961, 0.8039, 0.1961}{(+12.0)}} & \textbf{47.5 \textcolor[rgb]{0.1961, 0.8039, 0.1961}{(+13.3)}} & \textbf{51.3 \textcolor[rgb]{0.1961, 0.8039, 0.1961}{(+21.9)}}  & \textbf{47.9 \textcolor[rgb]{0.1961, 0.8039, 0.1961}{(+20.5)}} & \textbf{21.9 \textcolor[rgb]{0.1961, 0.8039, 0.1961}{(+8.6)}} & \textbf{40.4 \textcolor[rgb]{0.1961, 0.8039, 0.1961}{(+17.0)}}  \\
    \bottomrule
  \end{tabular}}
   
  \label{tab:pid}
  
\end{table*}

In this paper, we conduct experiments on two popular X-ray datasets, i.e., OPIXray \cite{wei2020occluded} and PIDray \cite{zhang2023pidray}.
%We also conduct experiments on the MS COCO 2017 dataset [XX] to validate the generalization ability of our method. 
OPIXray contains 8,885 images with 5 categories of prohibited items (i.e., different types of cutters).
Following \cite{wei2020occluded}, we use 7,109 images for training and 1,776 images for testing. We report mAP@.5 and mAP@[.5, .95] as the evaluation metrics. 
PIDray contains 29,457 images for training and 18,220 images for testing, covering 12 different categories. The images in the test set are further divided into 3 subsets (i.e., easy, hard, and hidden) according to their detection difficulty.
Following \cite{zhang2023pidray}, we use 29,457 images for training and 18,220 images for testing.
We report mAP@[.5, .95] as the evaluation metric.
%, and use mean average precision (mAP) as the evaluation metric to test the performance of the model in three different subsets.

To validate the generalization ability of our method to common object detection, we also conduct experiments on MS-COCO \cite{lin2014microsoft}.  MS-COCO is a public common object detection dataset, which contains more than 135k images for training and 5k images for testing, covering 80 different categories.
{Following \cite{li2020towards}, we use $train2017$ as training data, and report mAP@.5 and mAP@[.5, .95] on $val2017$.}
%we report the standard COCO-style average precision (AP) with IoU thresholds of 0.5:0.95 (AP) and 0.5 (AP$_{50}$) as evaluation metrics.

%All our experiments are conducted on the above three datasets and their noisy versions (i.e., OPIXray-Noise, PIDray-Noise, and MS COCO-Noise).

%\subsection{Artificial Noise Generation}

% \begin{table}[!t]\scriptsize
%   \centering
%   \caption{Bounding box Noise rate Estimation of OPIXray Dataset.}
%   \setlength{\tabcolsep}{0.4mm}{
%   \begin{tabular}{lcccccc}
%     \toprule
%     \multicolumn{1}{c}{Category} &  Scissor & Utility Knife & Multi-Tool Knife & Straight Knife & Folding Knife & Total\\
%     \midrule
%     Total Samples & 1494 & 1635 & 1612 & 809 & 1589 & 7139 \\
%     Noise Samples  &  62 & 60 & 51 & 34 & 20 & 227\\
%     Noise Rate  &  4.15\% & 3.67\% & 3.16\% & 4.20\% & 1.26\% & 3.18\%\\                             
%     \bottomrule
%   \end{tabular}}

%   \label{tab:small-loss}
  
% \end{table}

\begin{table}[!t]\scriptsize
  \centering
  \caption{{Bounding box noise rate estimation of the OPIXray Dataset.}}
  \setlength{\tabcolsep}{3.4mm}{
  \begin{tabular}{lccc}
    \toprule
    \multicolumn{1}{c}{Category} & Total Samples  & Noise Samples & Noise Rate \\
    \midrule
    Scissor & 1494 & 73 & 4.89\% \\
    Utility Knife  &  1635 & 70 & 4.28\%\\
    Multi-Tool Knife  &  1612 & 81 & 5.02\%\\
    Straight Knife  &  809 & 34 & 4.20\%\\
    Folding Knife  &  1589 & 33 & 2.08\%\\
    Total  &  7139 & 291 & 4.08\%\\                             
    \bottomrule
  \end{tabular}}

  \label{tab:estimation}
  
\end{table}

\subsection{Noise Rate Estimation}\label{sec:estimation}
{Our research has shown variability in noise rates across different X-ray datasets. Specifically, while some X-ray datasets (such as the PIDray dataset) exhibit minimal noisy annotations, we identify that some X-ray datasets (such as the OPIXray dataset) contain a number of noisy annotations (including both category noise and bounding box noise). In these datasets, many bounding box annotations are larger than the actual position of prohibited items while the category labels of some prohibited items are mislabeled due to the great similarity between some prohibited items.}

{We conduct a quantitative analysis of the noise rate on the OPIXray dataset. Specifically, for bounding box noise, we first train a model on the original dataset by treating all the categories as one category. In this  way, the influence of category noise is removed. Then, we compare the detection results with the ground-truth labels and filter out samples whose IoUs are less than 0.70. These samples are manually checked to identify noisy-labeled samples.
%use the original dataset to train a model to make a preliminary estimate of the noise rate. To improve the accuracy of bounding box prediction, we treat all categories as the same category during training (that is, only detect objects without considering categories), then compare the detection results with the ground-truth labels and filter out samples with IoU less than 0.7. These samples are then further filtered manually to find the real noise samples. 
For category noise, we randomly select a certain number of samples (500 samples in total) in the dataset and manually check whether they are labeled incorrectly. Based on the above steps, the category noise rate in the OPIXray dataset is estimated as about 5\%. The bounding box noise rate in the OPIXray dataset is estimated as about 4\%. We also estimate the bounding box noise rate for each class in the OPIXray dataset in Table \ref{tab:estimation}. 
The above analysis validates the existence of noisy annotations in some X-ray datasets.}

%\textcolor{red}{
%Although the noise rate in the dataset is relatively low, we believe our proposed method remains highly practical. First, obtaining high-quality human annotations is both costly and time-consuming \cite{su2012crowdsourcing, kuznetsova2020open}. To reduce annotation costs, some datasets rely on fewer annotators or use machine-generated labels. However, these approaches often introduce label noise (e.g., incorrect classes) and bounding box noise (e.g., inaccurate locations), both of which can negatively impact the learning process. Therefore, our method holds significant practical value in addressing these challenges.}

\subsection{Implementation Details}\label{sec:detail}
To effectively evaluate the performance of our method under noisy annotations, we introduce different types of noise to the original dataset. %This was achieved by artificially injecting generated noise into the original datasets.
For category noise, we randomly replace the original category label with another category label, with a replacement probability of $P_c$. 
For bounding box noise, we randomly perturb the original bounding box with a probability of $P_b$. Specifically, for a bounding box with coordinates $(x, y, w, h)$, we randomly perturb the coordinates with a probability of $P_b$ by shifting and scaling the box as follow:
\begin{equation}
   \begin{aligned}
   \widetilde{x} &= x + \Delta_{x} \times w, \\
   \widetilde{y} &= y + \Delta_{y} \times h, \\
   \widetilde{w} &= w \times (1 + \Delta_{w}), \\
    \widetilde{h} &= h \times (1 + \Delta_{h}), \\
    \end{aligned}
\end{equation}
where $\Delta_{x}$, $\Delta_{y}$, $\Delta_{w}$, and $\Delta_{h}$ are randomly sampled from a uniform distribution $U(-\delta, \delta)$ ($\delta$ is the perturbation level).
We set $\delta$ to 0.3 in all experiments.

For all the datasets, we adopt Faster R-CNN (FRCNN) \cite{ren2015faster} with ResNet-50 as the backbone network.
{The backbone is 
initialized with the weights pretrained on ImageNet \cite{krizhevsky2012imagenet}.}
%, and FPN is employed as a default. 
The whole network is optimized by the stochastic gradient descent (SGD) algorithm with a momentum of 0.9 and a weight decay of 0.0001. The batch size is set to 2. The initial learning
rate is set to 0.005 and decreased by a factor of 10 at the 17th and 21st epochs. The total number of training epochs is 24. The number of item patches $K$ for mixing is set to 2. 
We apply Mix-Paste to the training set with a probability of 0.6. {%We train the model using the mmdetection framework. 
We only employ
the random flip to all the comparison methods.}
All the competing methods are trained on a machine with an NVIDIA RTX 3090 GPU.

\subsection{Experiments on the X-Ray Datasets}\label{sec:exp-xray}
To verify the effectiveness of our method, we perform experiments on two X-ray datasets with different levels of noise rates for both category noise and bounding box noise.
%In our experiments, the noise rate refers to the combination of category noise and bounding box noise. For instance, a noise rate of 0.5 indicates that both the category noise rate and the bounding box noise rate are set to 0.5.
{We compare our method with several state-of-the-art methods, including the baseline method (FRCNN \cite{ren2015faster}), prohibited item detection methods (LIM \cite{tao2021towards}, SDANet \cite{zhang2023pidray},  {and GADet \cite{10322651}}), and learning with noisy annotations methods (SCE \cite{wang2019symmetric}, LNCIS \cite{yang2020learning}, and OA-MIL \cite{liu2022robust}).}
The results are shown in Table \ref{tab:opi} and Table \ref{tab:pid}. 

For the \textbf{OPIXray} dataset, we can observe that the noise-robust loss function-based method SCE does not perform well.
Compared with the baseline, SCE only gives marginal performance improvements at low noise rates. Moreover, when the noise rates are large, the performance obtained by SCE is even lower than that obtained by the baseline method.
The performance degradation of SCE can be attributed to its limited ability to handle the bounding box noise. In other words, when both the bounding box noise rate and the category noise rate are high, the performance of SCE is severely affected.
The X-ray prohibited item detector LIM shows relatively good anti-noise ability. At some noise rates, it even performs better than LNCIS, a method specifically designed to deal with object detection noise. This is because LIM can filter out irrelevant noisy information in features, making it less prone to overfit the noise. 
{The X-ray prohibited item detector GADet also demonstrates good performance in terms of mAP@[.5,.95].  
This can be attributed to its IoU-aware label assignment strategy, which selects high-quality and precise positive samples while ignoring potentially noisy low-quality predictions.}
Among all the competing methods, our Mix-Paste method achieves the best results at all noise rates.
Specifically, our method achieves 81.8\% mAP@.5 and 33.7\% mAP@[.5, .95] (at the noise rates of  $P_c = 60\%$ and $P_b=60\%$), which is 25.1\% and 15.3\% higher than the baseline, respectively.

For the \textbf{PIDray} dataset, the X-ray prohibited item detector LIM {and GADet} shows good performance. 
LNCIS can also alleviate the negative influence of noisy annotations to a certain extent. However, it exhibits only marginal performance improvements at high noise rates.
At some high noise rates, the performance obtained by OA-MIL is inferior to the baseline, indicating its instability.
Compared with the other competing methods, our method consistently gives the best results across all noise rates.
Specifically, at the noise rates of $P_c = 60\%$ and $P_b=60\%$, our method achieves a mAP@[.5, .95] of 51.3\%, 47.9\%, and 21.9\% in the easy, hard, and hidden test sets, which is 21.9\%, 20.5\%, and 8.6\% higher than the baseline, respectively.

The above results show that our method can greatly improve the performance of the model at both low and high noise rates and enhance the robustness of the model.

\begin{table}\scriptsize
  \caption{Comparison Results (\%) on the MS-COCO Dataset. $P_c$ and $P_b$ Denote the Category Noise Rate and Bounding Box Noise Rate, Respectively.}
  \centering
  \setlength{\tabcolsep}{0.5mm}{
  \begin{tabular}{lccccc}
    \toprule
    \multicolumn{1}{c}{\multirow{2}{*}{Method}} & \multicolumn{2}{c}{$P_c=40\% \quad P_b=40\%$}  & \multicolumn{2}{c}{$P_c=60\% \quad P_b=60\%$} \\
    \cmidrule(lr){2-3} \cmidrule(lr){4-5}
     & mAP@.5 & mAP@[.5, .95] & mAP@.5 & mAP@[.5, .95] \\
     \midrule
     FRCNN (PAMI, '17) \cite{ren2015faster} & 50.4 (+0.0) & 30.2 (+0.0) & 43.2 (+0.0) & 23.3 (+0.0) \\
     SCE (ICCV, '19) \cite{wang2019symmetric} & 44.6 \textcolor{red}{(-5.8)} & 27.1 \textcolor{red}{(-3.1)} & 36.8 \textcolor{red}{(-6.4)} & 20.5 \textcolor{red}{(-2.8)} \\
     LNCIS (ECCV, '20) \cite{yang2020learning} & 50.9 \textcolor[rgb]{0.1961, 0.8039, 0.1961}{(+0.5)} & 30.9 \textcolor[rgb]{0.1961, 0.8039, 0.1961}{(+0.7)} & 44.2 \textcolor[rgb]{0.1961, 0.8039, 0.1961}{(+1.0)} & 24.8 \textcolor[rgb]{0.1961, 0.8039, 0.1961}{(+1.5)} \\
     OA-MIL (ECCV, '22) \cite{liu2022robust} & 47.4 \textcolor{red}{(-3.0)} & 28.2 \textcolor{red}{(-2.0)} & 43.8 \textcolor[rgb]{0.1961, 0.8039, 0.1961}{(+0.6)} & 24.7 \textcolor[rgb]{0.1961, 0.8039, 0.1961}{(+1.4)} \\
     Ours & \textbf{51.2 \textcolor[rgb]{0.1961, 0.8039, 0.1961}{(+0.8)}} & \textbf{31.5 \textcolor[rgb]{0.1961, 0.8039, 0.1961}{(+1.3)}} & \textbf{45.3 \textcolor[rgb]{0.1961, 0.8039, 0.1961}{(+2.1)}} & \textbf{26.2 \textcolor[rgb]{0.1961, 0.8039, 0.1961}{(+2.9)}} \\
    \bottomrule
  \end{tabular}}
  
  \label{tab:coco}
\end{table}

\subsection{Experiments on the MS-COCO Dataset}\label{sec:exp-coco}
{Our method is mainly designed 
for prohibited item detection under noisy annotations by considering the characteristics of X-ray images, where item overlapping is ubiquitous. Interestingly, object overlapping also exists in some natural images for the common object detection task. Hence, the idea of increasing the probability of target objects in the generated images by fusing the same category label can be also applied to common object detection.}

{To evaluate the generalizability of our proposed method, we conduct experiments on the widely used common object detection dataset, the MS-COCO dataset.
The evaluation results are shown in Table \ref{tab:coco}. From Table \ref{tab:coco}, we observe that SCE  performs poorly. LNCIS shows moderate performance improvements over the baseline method (the vanilla FRCNN). %, unfortunately, this enhancement is relatively modest.
The performance of OA-MIL is unstable, and its performance is even worse than the baseline at some noise rates (e.g., $P_c=40\%$ and $P_b=40\%$).
Compared with the other competing methods, our method gives better performance at different noise rates. 
%demonstrates the most noticeable enhancement in model performance under noise scenarios.
Specifically, at the noise rates of $P_c = 60\%$ and $P_b=60\%$, our method achieves 45.3\% mAP@.5 and 26.2\% mAP@[.5, .95], which is 2.1\% and 2.9\% higher than the baseline, respectively.
These results demonstrate the effectiveness of our method on the noisy MS-COCO dataset, indicating the robustness and broad applicability of our method to data beyond X-ray images.
}

\subsection{Ablation Studies}\label{sec:ablation}

We conduct ablation studies to study the effectiveness of each component in our method. Unless otherwise specified, the noise rates are set to $P_c$ = 60\% and $P_b$ = 60\% and the LLS strategy is not used (we focus on the evaluation of Mix-Paste).
The OPIXray dataset is used.
\begin{table}\scriptsize
  
  \caption{Ablation Study Results (\%) on the Key Components of Our Method on the OPIXray Dataset.}
  \centering
  \setlength{\tabcolsep}{1.5mm}{
  \begin{tabular}{lccccc}
    \toprule
    \multicolumn{1}{c}{Method} & \multicolumn{1}{c}{Mix-Paste} & \multicolumn{1}{c}{LLS} & mAP@.5 & mAP@[.5, .95]\\
    \midrule
    FRCNN (PAMI, '17) \cite{ren2015faster} &  &  & 56.7 & 18.4\\
    \multicolumn{1}{c}{\multirow{2}{*}{Ours}} & \multicolumn{1}{c}{\Checkmark} & & 80.3 & 31.5\\
                                              & \multicolumn{1}{c}{\Checkmark} & \multicolumn{1}{c}{\Checkmark} & 81.8 & 33.7\\
    % & & \\
    % Ours & Makes one's heart Frob\\
    \bottomrule
  \end{tabular}}
  \label{tab:abu}
\end{table}

\begin{figure}
  \centering
  \subfloat[]{
    \includegraphics[width=0.23\textwidth]{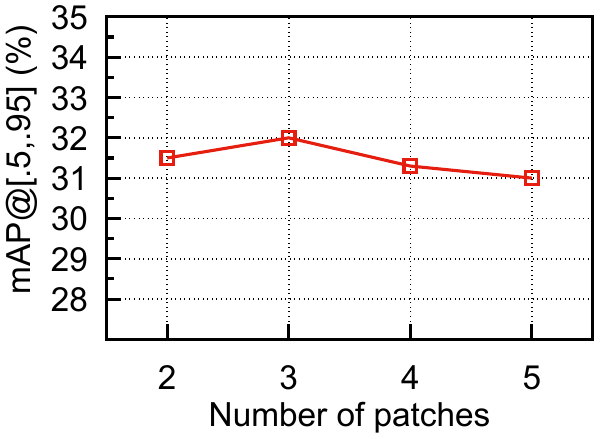}
    \label{fig:number}}
  \subfloat[]{
    \includegraphics[width=0.23\textwidth]{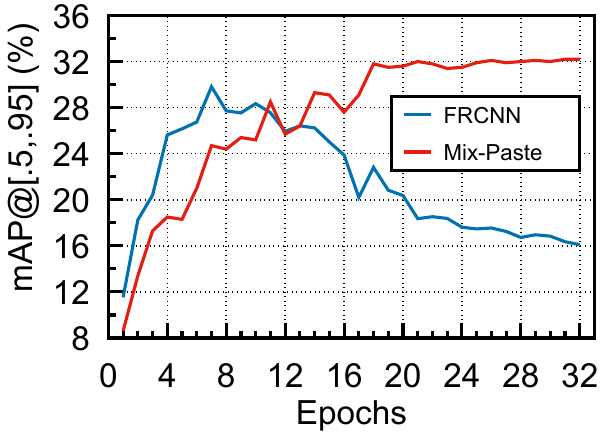}
    \label{fig:schedule}}
   \caption{(a) Ablation study results on the influence of the number of patches for Mix-Paste on the OPIXray dataset. (b) {Training curve comparison between our Mix-Paste and the baseline method on the OPIXray dataset.} }
   \label{fig:3}
\end{figure}

\noindent  \textbf{Effectiveness of Mix-Paste and LLS.}
The ablation study results on the key components of our method are shown in Table \ref{tab:abu}. 
We can see that the performance of our method with only Mix-Paste is better than the baseline by 13.1\% mAP@[.5, .95]. This demonstrates the effectiveness of Mix-Paste, which mixes multiple item patches with the same category label to alleviate the influence of noisy annotations. After further applying the LLS strategy, the performance of our method is further improved by 2.2\%, verifying the importance of the LLS strategy, which ignores the potentially positive predictions during the mixing process. 

\noindent \textbf{Influence of the Number of Patches $K$.}
We investigate the influence of the number of patches $K$ used in Mix-Paste, as shown in Fig. \ref{fig:3}(a). Our method gives a good performance when the values of $K$ are set to 2 and 3. However, when the value of $K$ becomes large (e.g., 4 or 5), the performance obtained by our method decreases. When more patches are mixed,  the likelihood of capturing the correct prohibited item is increased. However, such a way also raises the probability of introducing additional prohibited items. As a result, excessive patch mixing can {negatively} influence the extraction of relevant information in the original patch,  hindering the learning of correct prohibited items. 

\noindent  \textbf{Training Curve.}
To verify whether our Mix-Paste can help alleviate the overfitting of noise during model training, we plot the training curve in Fig. \ref{fig:3}(b). 
We can see that the performance obtained by the baseline model increases at the early training stage but gradually decreases at the later training stage. In contrast, the performance obtained by our method is relatively stable at the later training stage. This demonstrates that our method can effectively alleviate the overfitting of the model to noise during training.

\begin{table}[!t]\footnotesize
  
  \caption{Ablation Study Results (\%) on the Influence of the Edge Smoothing Mask on the OPIXray Dataset {without introducing any additional noise}. {Only Mix-Paste Is Used in This Experiment.}}
  \centering
  \setlength{\tabcolsep}{3mm}{
  \begin{tabular}{ccccccc}
    \toprule
    
     Method & Linear Combination & Edge Smoothing Mask \\
    % \multicolumn{1}{c}{Method} &  mAP@.5 & mAP@[.5, .95]\\
    \midrule
    % numk &  56.7 & 18.4\\
    % mAP@[.5,.95] & 18.4 & 32.6 & 32.8 & 32.9 & 30.8 & 0.4 \\
    mAP@[.5,.95] & 31.3 & 31.5 \\
                                              
    % & & \\
    % Ours & Makes one's heart Frob\\
    \bottomrule
  \end{tabular}}

  \label{tab:mask}
  
\end{table}

\begin{table}[!t]
  
  \caption{Comparison of Training Time, Inference Time and Performance (\%) on the Original OPIXray Dataset without Introducing Additional Noise.}
  \centering
  \scriptsize
  \setlength{\tabcolsep}{1.3mm}{
  \begin{tabular}{lcccc}
    \toprule
    \multicolumn{1}{c}{\multirow{2}{*}{Method}}  & \multicolumn{1}{c}{\multirow{2}{*}{mAP@.5}} & \multicolumn{1}{c}{\multirow{2}{*}{mAP@[.5,.95]}}&  Training & Inference \\
    & & & Time & Time(fps)\\
    \midrule
    FRCNN (PAMI, '17) \cite{ren2015faster}   & 86.1 &  36.9&  4h 27min & 18.7 \\
    LIM (ICCV, '21) \cite{tao2021towards}   & 88.6 & 38.9 & 19h 18min& 7.3 \\
    SDANet (IJCV, '23) \cite{zhang2023pidray}  & 88.1 & 38.3& 6h 20min & 16.2 \\
    SCE (ICCV, '19) \cite{wang2019symmetric}  & 85.9 & 36.7& 4h 28min & 19.3 \\
    LNCIS (ECCV, '20) \cite{yang2020learning}  & 88.1 & 37.6& 4h 30min & 19.3 \\
    OA-MIL (ECCV, '22) \cite{liu2022robust}  & 87.3 & 37.4 & 4h 33min& 19.4 \\
    {Mix-Paste + LLS}  & \textbf{90.1} & \textbf{40.3}& 4h 30min & 19.0 \\

    \bottomrule
  \end{tabular}}
  
  \label{tab:original}
  
\end{table}

\begin{table}\scriptsize
  
  \caption{{Ablation Study Results (\%) on the effectiveness of the patch mixing strategy. Random-Mix denotes the method that selects two random item patches for the mixing operation. The noise rates are set to $P_c$ = 60\% and $P_b$ = 60\%}}
  \centering
  \setlength{\tabcolsep}{5mm}{
  \begin{tabular}{lcc}
    \toprule
    \multicolumn{1}{c}{Method} &  mAP@.5 & mAP@[.5, .95]\\
    \midrule
    FRCNN (PAMI, '17) \cite{ren2015faster} &  56.7 & 18.4\\
    Random-Mix  & 55.7 & 21.0\\
    Mix-Paste &   \textbf{80.3} & \textbf{31.5}\\

    \bottomrule
  \end{tabular}}
  \label{tab:randommix}
  
\end{table}

\noindent {\textbf{Effectiveness of the Patch Mixing Strategy.}
In Mix-Paste, 
we mix multiple item patches with the same category label to generate a mixed patch and paste it back into
the original image for data augmentation.
To show the effectiveness of our patch mixing strategy, we compare the performance between our strategy and a variant (which mixes multiple randomly selected item patches with different category labels).
The results are shown in Table \ref{tab:randommix}.}

{We can see that 
our patch mixing strategy achieves better performance than the variant. Mixing item patches with different category labels not only brings additional disturbances caused by different prohibited items, but also does not increase the probability of containing the correct prohibited item in the generated image, thereby decreasing the final performance. The above results further validate the superiority of our patch mixing strategy.}
%This is consistent with our ablation study results on the number of patches $K$ (our method gives a good performance
%when the values of $K$ are set to 2 and 3), where excessive
%patch mixing affects the learning of correct prohibited items.} 

\begin{figure*}[!t]
  \centering \includegraphics[width=14cm]{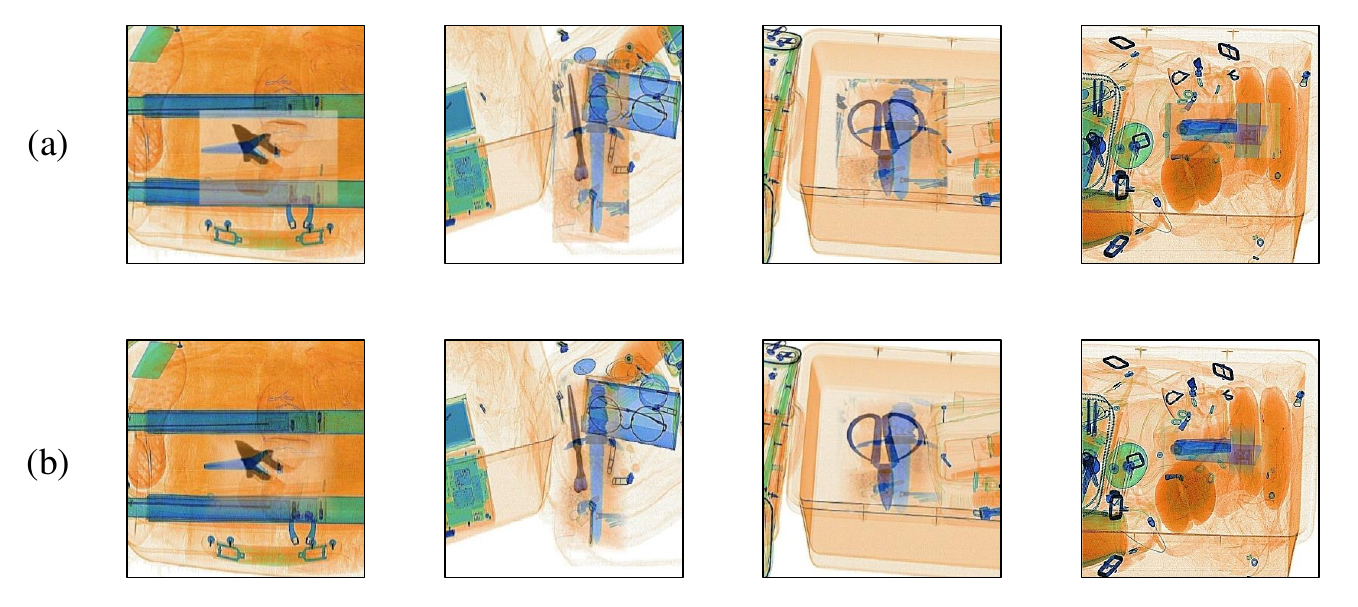}
  \caption{Examples of generated images on the OPIXray dataset. (a) The generated images with the linear combination. (b) The generated images with the edge smoothing mask.}
  \label{fig:mask}
\end{figure*}

\noindent \textbf{Effectiveness of the Edge Smoothing Mask.}
For our Mix-Paste, we use an edge smoothing mask to mix multiple item patches, generating images with more natural appearances (some generated images are illustrated in Fig. \ref{fig:mask}). We conduct experiments to investigate the influence of the edge smoothing mask. We compare the edge smoothing mask with the simple linear combination of multiple item patches (i.e., all the pixels in the edge smoothing mask are set to a fixed value). 
The results are given in Table \ref{tab:mask}. 
We can see that the performance obtained by our method with the edge smoothing mask is only slightly better than that with the linear combination. This can be ascribed to the fact that the model focuses on the appearance of the prohibited items, and thus it does not pay too much attention to the edges of the mixed patches during the learning process. 
This result also aligns with the principle of Occam's razor, where the linear combination indicates a simple fusion method.

\noindent \textbf{Effectiveness on the Original X-ray Dataset.}
We conduct experiments to investigate the effectiveness of our method on the original X-ray dataset. We test all the competing methods on the OPIXray dataset without introducing additional noise.
We also report the training time {and inference time} of different methods. 
The results are shown in Table \ref{tab:original}.
From Table \ref{tab:original}, we can observe that our method achieves better performance than the other competing methods. This is because the original dataset also contains a certain amount of noisy annotations, and our method can effectively alleviate the negative influence of noisy annotations during the model training.
Note that although the LIM and SDANet methods also outperform the baseline method (FRCNN), the training time {and inference time} of these methods is significantly longer.  
%Notably, the LIM method quadruples the training time when compared to baseline models.

\noindent \textbf{Effectiveness on Different Category Noise Rates and Bounding Box Noise Rates.}
We conduct experiments to investigate the effectiveness of our method on different category noise rates and bounding box noise rates on the OPIXray dataset. The results are shown in Table \ref{tab:different}.
From the results, we can see that OA-MIL is good at addressing category noise, and LNCIS works well on handling bounding box noise. Among all the competing methods, our method can effectively deal with both category noise and bounding box noise, and achieve the best results in all the cases. This proves that our method has a strong anti-noise ability.

\begin{table}[!t] \scriptsize
  
  \caption{Comparison Results (\%) of Different Category Noise Rates and Bounding Box Noise Rates on the OPIXray Dataset. $P_c$ and $P_b$ Denote the Category Noise Rate and the Bounding Box Noise Rate, Respectively. }
  \centering
  \setlength{\tabcolsep}{0.6mm}{
  \begin{tabular}{lcccc}
    \toprule
    \multicolumn{1}{c}{\multirow{2}{*}{Method}} &  \multicolumn{2}{c}{$P_c=20\% \quad P_b=40\%$} &\multicolumn{2}{c}{$P_c=40\% \quad P_b=20\%$}  \\
    \cmidrule(lr){2-3} \cmidrule(lr){4-5}  
      & \multicolumn{1}{c}{mAP@.5} & \multicolumn{1}{c}{mAP@[.5, .95]} & \multicolumn{1}{c}{mAP@.5} & \multicolumn{1}{c}{mAP@[.5, .95]}  \\
     \midrule
     FRCNN (PAMI, '17) \cite{ren2015faster}  & 78.4 (+0.0) & 28.9 (+0.0) & 72.7 (+0.0) & 28.5 (+0.0) \\
     LIM (ICCV, '21) \cite{tao2021towards}  & 82.0 \textcolor[rgb]{0.1961, 0.8039, 0.1961}{(+3.6)} & 32.2 \textcolor[rgb]{0.1961, 0.8039, 0.1961}{(+3.3)} & 81.5 \textcolor[rgb]{0.1961, 0.8039, 0.1961}{(+8.8)} & 32.7 \textcolor[rgb]{0.1961, 0.8039, 0.1961}{(+4.2)}\\
     SDANet (IJCV, '23) \cite{zhang2023pidray}  & 80.1 \textcolor[rgb]{0.1961, 0.8039, 0.1961}{(+1.7)} & 28.3 \textcolor{red}{(-0.6)} & 74.4 \textcolor[rgb]{0.1961, 0.8039, 0.1961}{(+1.7)} & 29.6 \textcolor[rgb]{0.1961, 0.8039, 0.1961}{(+1.1)} \\
     GADet (SENS J., '24) \cite{10322651} & 77.5 \textcolor{red}{(-0.9)} & 32.1 \textcolor[rgb]{0.1961, 0.8039, 0.1961}{(+3.1)} & 77.9 \textcolor[rgb]{0.1961, 0.8039, 0.1961}{(+5.2)} & 34.0 \textcolor[rgb]{0.1961, 0.8039, 0.1961}{(+5.5)} \\
     SCE (ICCV, '19) \cite{wang2019symmetric} & 76.7 \textcolor{red}{(-1.7)} & 27.3 \textcolor{red}{(-1.6)} & 75.4 \textcolor[rgb]{0.1961, 0.8039, 0.1961}{(+2.7)} & 29.8 \textcolor[rgb]{0.1961, 0.8039, 0.1961}{(+1.3)} \\
     LNCIS (ECCV, '20) \cite{yang2020learning} & 78.9 \textcolor[rgb]{0.1961, 0.8039, 0.1961}{(+0.5)} & 28.5 \textcolor{red}{(-0.4)} & 81.9 \textcolor[rgb]{0.1961, 0.8039, 0.1961}{(+9.2)} & 33.3 \textcolor[rgb]{0.1961, 0.8039, 0.1961}{(+4.8)}  \\
     OA-MIL (ECCV, '22) \cite{liu2022robust}  & 54.8 \textcolor[rgb]{0.1961, 0.8039, 0.1961}{(+3.8)} & 20.8 \textcolor[rgb]{0.1961, 0.8039, 0.1961}{(+2.4)} & 73.3 \textcolor[rgb]{0.1961, 0.8039, 0.1961}{(+0.6)} & 28.2 \textcolor{red}{(-0.3)}  \\
     {Mix-Paste + LLS} & \textbf{86.2 \textcolor[rgb]{0.1961, 0.8039, 0.1961}{(+7.8)}} & \textbf{36.9 \textcolor[rgb]{0.1961, 0.8039, 0.1961}{(+8.0)}} & \textbf{86.3 \textcolor[rgb]{0.1961, 0.8039, 0.1961}{(+13.6)}} & \textbf{37.7 \textcolor[rgb]{0.1961, 0.8039, 0.1961}{(+9.2)}} \\
    \bottomrule
  \end{tabular}}
  \label{tab:different}
\end{table}

\noindent  \textbf{Influence of the Probability of Applying Mix-Paste.}
We investigate the influence of the probability $p$ of applying Mix-Paste for augmentation during training. The results are shown in Table \ref{tab:probability}. 
We can see that when the value of $p$ is 0.6, our method can achieve the best performance. When the value of $p$ approaches 1, the performance obtained by our method decreases or even the training fails. 
This is because when $p$ is close to 1, all samples are artificially generated. Such a manner can lead to significant inconsistency between the training samples and the test samples, making the distribution of the training set greatly inconsistent with that of the test set. As a consequence, our method is unable to learn the true data distribution from the training set.
In this paper, we fix $p=0.6$ in all experiments. 
\begin{table}\footnotesize
  \centering
  \caption{Ablation Study Results (\%) on the Influence of the Probability of Applying Mix-Paste on the OPIXray Dataset. {Only Mix-Paste Is Used in This Experiment.}}
  \setlength{\tabcolsep}{3mm}{
  \begin{tabular}{ccccccc}
    \toprule
    Probability & 0 & 0.2 & 0.4 & 0.6 & 0.8 & 1\\
    % \multicolumn{1}{c}{Method} &  mAP@.5 & mAP@[.5, .95]\\
    \midrule
    % numk &  56.7 & 18.4\\
    % mAP@[.5,.95] & 18.4 & 32.6 & 32.8 & 32.9 & 30.8 & 0.4 \\
    mAP@[.5,.95] & 18.4 & 31.2 & 31.4 & 31.5 & 30.8 & 0.4 \\
                                              
    % & & \\
    % Ours & Makes one's heart Frob\\
    \bottomrule
  \end{tabular}}

  \label{tab:probability}
  
\end{table}

\begin{table}[!t]
  
  \caption{Comparison Results (\%) under Different Perturbation Levels on the OPIXray Dataset.}
  \centering
  \scriptsize
  \setlength{\tabcolsep}{0.5mm}{
  \begin{tabular}{lllll}
    \toprule
    \multicolumn{1}{c}{\multirow{3}{*}{Method}} &  \multicolumn{2}{c}{$\delta=0.3$} &\multicolumn{2}{c}{$\delta=0.5$}  \\
    &  \multicolumn{2}{c}{$P_c=60\% \quad P_b=60\%$} &\multicolumn{2}{c}{$P_c=60\% \quad P_b=60\%$} \\
    \cmidrule(lr){2-3} \cmidrule(lr){4-5}  
      & \multicolumn{1}{c}{mAP@.5} & \multicolumn{1}{c}{mAP@[.5, .95]} & \multicolumn{1}{c}{mAP@.5} & \multicolumn{1}{c}{mAP@[.5, .95]}  \\
    \midrule
    FRCNN (PAMI, '17) \cite{ren2015faster}  & 56.7 (+0.0) & 18.4 (+0.0) & 42.7 (+0.0) &  12.7 (+0.0) \\
    LIM (ICCV, '21) \cite{tao2021towards}  & 72.7 \textcolor[rgb]{0.1961, 0.8039, 0.1961}{(+15.0)} & 26.4 \textcolor[rgb]{0.1961, 0.8039, 0.1961}{(+8.0)} & 65.7 \textcolor[rgb]{0.1961, 0.8039, 0.1961}{(+23.0)} & 23.6 \textcolor[rgb]{0.1961, 0.8039, 0.1961}{(+10.9)} \\
    SDANet (IJCV, '23) \cite{zhang2023pidray} & 52.4 \textcolor{red}{(-4.3)} & 17.1 \textcolor{red}{(-1.3)} & 38.2 \textcolor{red}{(-4.5)}  & 10.7 \textcolor{red}{(-2.0)} \\
    GADet (SENS J., '24) \cite{10322651} & 69.7 \textcolor[rgb]{0.1961, 0.8039, 0.1961}{(+13.0)} & 27.9 \textcolor[rgb]{0.1961, 0.8039, 0.1961}{(+9.5)} & 59.6 \textcolor[rgb]{0.1961, 0.8039, 0.1961}{(+16.9)}  & 24.1 \textcolor[rgb]{0.1961, 0.8039, 0.1961}{(+7.4)} \\
    SCE (ICCV, '19) \cite{wang2019symmetric} & 48.6 \textcolor{red}{(-8.1)} & 16.1 \textcolor{red}{(-2.3)} & 35.0 \textcolor{red}{(-7.7)} & 10.5 \textcolor{red}{(-2.2)} \\
    LNCIS (ECCV, '20) \cite{yang2020learning} & 65.9 \textcolor[rgb]{0.1961, 0.8039, 0.1961}{(+9.2)} & 23.4 \textcolor[rgb]{0.1961, 0.8039, 0.1961}{(+5.0)} & 46.6 \textcolor[rgb]{0.1961, 0.8039, 0.1961}{(+3.9)} & 14.0 \textcolor[rgb]{0.1961, 0.8039, 0.1961}{(+1.3)} \\
    OA-MIL (ECCV, 
    22) \cite{liu2022robust}   & 56.4 \textcolor{red}{(-0.3)} & 21.6 \textcolor[rgb]{0.1961, 0.8039, 0.1961}{(+3.2)} & 44.9 \textcolor[rgb]{0.1961, 0.8039, 0.1961}{(+2.2)} & 15.0 \textcolor[rgb]{0.1961, 0.8039, 0.1961}{(+2.3)} \\
   {Mix-Paste+LLS} & \textbf{81.8 \textcolor[rgb]{0.1961, 0.8039, 0.1961}{(+25.1)}} & \textbf{33.7 \textcolor[rgb]{0.1961, 0.8039, 0.1961}{(+15.3)}} & \textbf{76.1 \textcolor[rgb]{0.1961, 0.8039, 0.1961}{(+33.4)}} & \textbf{29.9 \textcolor[rgb]{0.1961, 0.8039, 0.1961}{(+17.2)}} \\
                                              
    % & & \\3.3+21.8
    % Ours & Makes one's heart Frob\\
    \bottomrule
  \end{tabular}}
  \label{tab:perturbation}
  
\end{table}

\begin{table}[!t]\scriptsize
  \centering
  \caption{{The mAP@[.5, .95] (\%) and mAP@.5 Obtained by Different Detectors on the OPIXray Dataset with noise rate of $P_c$ = 60\% and $P_b$ = 60\%. {Only Mix-Paste Is Used in This Experiment.}}}
  \setlength{\tabcolsep}{0.5mm}{
  \begin{tabular}{lcccc}
    \toprule
    \multicolumn{1}{c}{\multirow{2}{*}{Method}} & \multicolumn{2}{c}{Original} & \multicolumn{2}{c}{+Mix-Paste}\\
     & \multicolumn{1}{c}{mAP@.5} & \multicolumn{1}{c}{mAP@[.5, .95]} & \multicolumn{1}{c}{mAP@.5} & \multicolumn{1}{c}{mAP@[.5, .95]}\\
    \midrule
    FRCNN (PAMI, '17) \cite{ren2015faster} & 56.7 & 18.4 & \textbf{80.3} & \textbf{31.5} \\
    RetinaNet (ICCV, '17) \cite{lin2017focal} & 56.1 & 20.8 & 61.7 & 26.1\\
    Cascade RCNN (CVPR, '18) \cite{cai2018cascade} & 47.6 & 16.0 & 78.6 & 31.6\\
    ATSS (CVPR, '20) \cite{zhang2020bridging} & 55.5 & 18.7 & 68.2 & 28.2 \\
    LIM (ICCV, '21) \cite{tao2021towards} & \textbf{72.7} & \textbf{26.4} & 78.1 &  31.1 \\      
    SDANet (IJCV, '23) \cite{zhang2023pidray} & 52.4 & 17.1 & 77.8 & 31.6  \\
                              
    % & & \\
    % Ours & Makes one's heart Frob\\
    \bottomrule
  \end{tabular}}

  \label{tab:detector}
  
\end{table}

\begin{figure}[!t]
  \centering
  \subfloat[]{
    \includegraphics[width=0.23\textwidth]{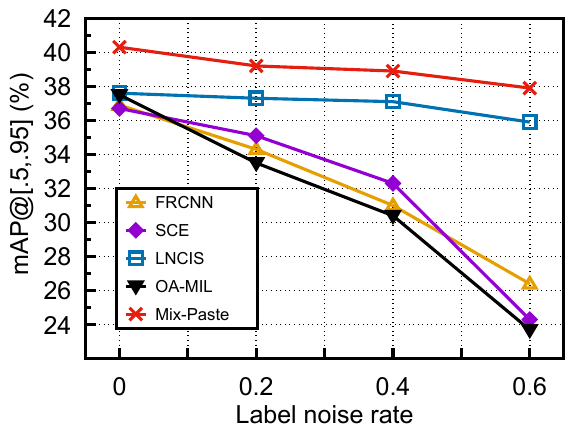}}
  \subfloat[]{
    \centering
    \includegraphics[width=0.23\textwidth]{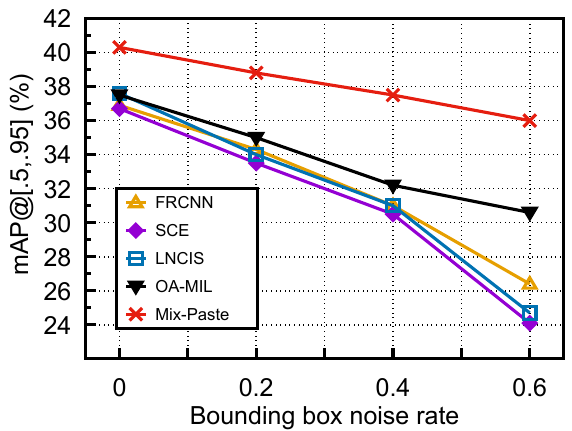}}
   \caption{Performance comparison between Mix-Paste and the competing methods at different noise rates under (a) category noise and (b) bounding box noise on the OPIXray dataset.}
   \label{fig:noisetype}
\end{figure}

\noindent \textbf{Influence of the Perturbation Level.}
We conduct experiments to investigate the influence of perturbation level (which is used to generate bounding box noise). We test our method under two perturbation levels (i.e., $\delta=0.5$ and $\delta=0.3$) on the OPIXray dataset. The results are shown in Table \ref{tab:perturbation}.
We can see that even when the perturbation level is large ($\delta=0.5$), our method can still achieve the best results, while other methods suffer from a serious performance decline. 
This shows that our method has a strong anti-noise ability in the case of high perturbation levels.

\noindent \textbf{Influence of Different Detectors.}
Our Mix-Paste, which is a data augmentation method, can be applied to different object detectors. 
{We evaluate the performance obtained by our method on different detectors, including two-stage detectors (FRCNN \cite{ren2015faster} and Cascade RCNN \cite{cai2018cascade}), one-stage detectors (RetinaNet \cite{lin2017focal} and ATSS \cite{zhang2020bridging}), and X-ray prohibited item detectors (SDANet \cite{zhang2023pidray} and LIM \cite{tao2021towards}). {For all the detectors, we use the default hyper parameters in the  MMDetection framework\cite{chen2019mmdetection} for training.} The results are shown in Table \ref{tab:detector}.}
We can observe that all the detectors with our method outperform those without our method. These results show the superiority of our proposed Mix-Paste. {Note that the experimental settings in Table \ref{tab:detector} and Table \ref{tab:original} are different. In Table \ref{tab:detector}, the results are obtained on the noisy OPIXray dataset under the settings that both the bounding box
noise rate and the label noise rate are 60\%. Hence, the mAP obtained by FRCNN is
low (18.4\%). In Table \ref{tab:original}, the results are obtained on the original OPIXray dataset without introducing any synthetic noise. Hence, the
mAP obtained by FRCNN in Table \ref{tab:original} is higher than that in Table \ref{tab:detector}.}

\noindent  \textbf{Robustness to Different Types of Noise.}
We investigate the robustness of our method at different noise rates under two types of noise, including category noise and bounding box noise. The results are shown in Fig. \ref{fig:noisetype}. For category noise, it is evident that both SCE and OA-MIL struggle to mitigate the adverse influence of category noise on the model. In contrast, both LNCIS and our method show great effectiveness in dealing with category noise. Notably, our method exhibits the best performance at different noise rates, demonstrating its effectiveness in handling category noise.
For bounding box noise, our method outperforms the baseline method by a large margin at the high bounding box noise rates. Moreover, the performance obtained by some competing methods (such as SCE and LNCIS) shows a significant performance decline at high bounding box noise rates.  
This further validates the superiority of our method in handling bounding box noise.

\begin{figure*}[!t]
  \centering
  \includegraphics[width=14cm]{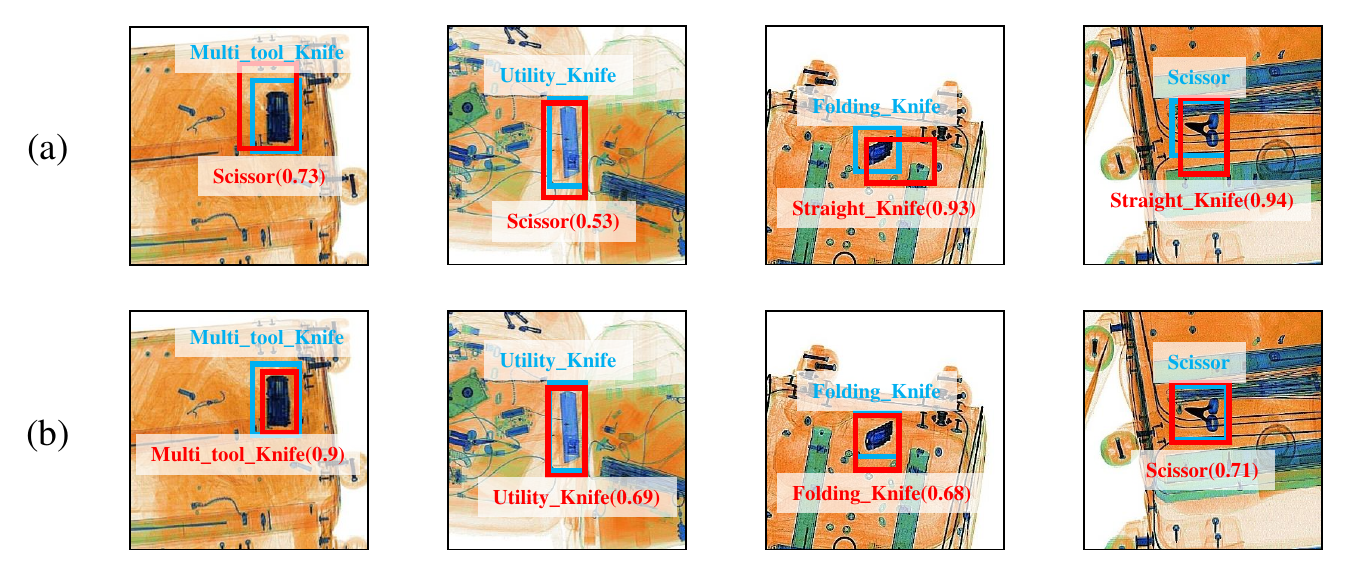}
  \caption{Some detection results on the OPIXray dataset. (a) The detection results obtained by the baseline method (FRCNN). (b) The detection results obtained by our method. The model is trained under the noise rates of $P_l=60\%$ and $P_b=60\%$. The blue annotation and bounding box denote the ground-truth annotation and bounding box, and the red annotation and bounding box denote the detection result.}
  \label{fig:results}
\end{figure*}

\begin{table}\scriptsize
  
  \caption{Comparison Results (\%) against Some Popular Data Augmentation Methods on the OPIXray Dataset.}
  \centering
  \setlength{\tabcolsep}{4mm}{
  \begin{tabular}{lcc}
    \toprule
    \multicolumn{1}{c}{Method} &  mAP@.5 & mAP@[.5, .95]\\
    \midrule
    FRCNN (PAMI, '17) \cite{ren2015faster} &  56.7 & 18.4\\
    Mix-Up (arXiv, '17) \cite{zhang2017mixup} & 46.2 & 15.5\\
    Cutout (arXiv, '17) \cite{devries2017improved} & 58.7 & 21.0 \\
    Mosaic (arXiv, '20) \cite{bochkovskiy2020yolov4} & 58.7 & 19.9 \\
    {Color jitter}
     & \multicolumn{1}{c}{\multirow{2}{*}{57.6}} & \multicolumn{1}{c}{\multirow{2}{*}{19.4}} \\
    (Commun. ACM, '17) \cite{krizhevsky2012imagenet} & &\\
    Blur & 61.7 & 21.8  \\
    Mix-Paste &  \textbf{80.3} & \textbf{31.5}\\

    \bottomrule
  \end{tabular}}
  
  \label{tab:mixup}
  
\end{table}

\noindent  \textbf{Comparison against  Popular Data Augmentation Methods.}
%Our method is similar to Mix-Up in that both methods combine samples to generate new training samples. However, the motivations behind the two methods are distinctly different. 
Table \ref{tab:mixup} shows the comparison results between our method and some popular data augmentation methods. We observe that the performance obtained by Mix-Up is even lower than that obtained by the baseline method. Our method is specifically designed to address the challenge of noisy annotations while Mix-Up does not consider such a challenge.
    In fact, the mixing process in Mix-Up cannot reduce the noise, and can significantly introduce additional noisy samples, leading to model overfitting to noisy annotations. 
    Moreover, our method greatly outperforms other competing methods (such as Mosaic, Cutout and Color jitter), showing its effectiveness.

\begin{table}[!t]\scriptsize
  \centering
  \caption{Comparison Results (\%) against the Small-Loss Criterion on the OPIXray Dataset.}
  \setlength{\tabcolsep}{5.4mm}{
  \begin{tabular}{lcc}
    \toprule
    \multicolumn{1}{c}{Method} &  mAP@.5 & mAP@[.5, .95]\\
    \midrule
    Mix-Paste & 80.3 & 31.5 \\
    Mix-Paste+Small-Loss V1  &  64.2 & 23.6\\
    Mix-Paste+Small-Loss V2  & 62.8 & 23.1\\
    Mix-Paste+Small-Loss V3  & 62.4 & 21.0 \\
    Mix-Paste+LLS & \textbf{81.8} & \textbf{33.7}\\
                                              
    \bottomrule
  \end{tabular}}

  \label{tab:small-loss}
  
\end{table}

\noindent \textbf{Comparison against the Small-Loss Criterion.}
%We conduct experiments to investigate the effectiveness of our method against small-loss. 
We compare our LLS with the small-loss criterion. 
The small-loss criterion is a widely used method in label noise learning for the image classification task.
Due to the difference between the object detection task and the image classification task, 
we implement three different versions of the small-loss criterion. The first version  (Small-Loss V1) is that we select samples with small losses from all classification losses as clean samples and add them to the loss calculation.
The second version (Small-Loss V2) is that we divide all classification results into two categories: positive predictions (whose IoUs between predicted boxes and the ground truth are greater than a threshold) and negative predictions (whose IoUs between predicted boxes and the ground truth are less than a threshold), and select a certain proportion of small-loss predictions from the two parts and add them to the loss calculation.
{The third version (Small-Loss V3) is that we select a certain proportion of samples with small losses from the total loss ({both the classification loss and the regression loss}) as clean samples and add them to the loss calculation.}
For a fair comparison, both the small-loss criterion and our LLS are based on our Mix-Paste.
The results are shown in Table \ref{tab:small-loss}. {The clean sample proportion in the small-loss criterion is set to $1-\tau \cdot min(T/5, 1)$ as done in \cite{han2018co}, where $\tau$ is the noise rate (we set it to 0.6) and $T$ is the training epoch. }

We can see that the three versions of the small-loss criterion fail to improve the performance of Mix-Paste. On the contrary, our LLS can be effectively combined with Mix-Paste to improve the performance, which validates the importance of LLS in learning with noisy annotations.

% \begin{table}\scriptsize
  
%   \caption{Ablation Study Results (\%) on the Influence of the Probability of bounding box noise distribution on the OPIXray Dataset. }
%   \centering
%   \setlength{\tabcolsep}{6mm}{
%   \begin{tabular}{lcc}
%     \toprule
%     \multicolumn{1}{c}{Method} &  mAP@.5 & mAP@[.5, .95]\\
%     \midrule
%     FRCNN \cite{ren2015faster} &  39.1 & 10.4\\
%     LIM \cite{zhang2017mixup} & 60.4 & 19.3\\
%     SDANet \cite{bochkovskiy2020yolov4} & 34.8 & 8.9 \\
%     % SCE \cite{devries2017improved} & 58.7 & 21.0 \\
%     LNCIS\cite{krizhevsky2012imagenet} & 49.0 & 16.5 \\
%     OA-MIL & 72.1 & 25.7  \\
%     Ours &   \textbf{80.3} & \textbf{31.5}\\

%     \bottomrule
%   \end{tabular}}
  
%   \label{tab:mixup}
% \end{table}

\begin{table}[!t] \scriptsize
  
  \caption{{Ablation Study Results (\%) on the Influence of the bounding box noise distribution on the OPIXray Dataset. We add noise to the bounding box using the Gaussian distribution.}}
  \centering
  \setlength{\tabcolsep}{0.5mm}{
  \begin{tabular}{lllll}
    \toprule
    \multicolumn{1}{c}{\multirow{3}{*}{Method}} &  \multicolumn{2}{c}{$P_c=60\%$} &\multicolumn{2}{c}{$P_c=60\%$}  \\
    % \cmidrule(lr){2-3} \cmidrule(lr){4-5} 
    & \multicolumn{2}{c}{$\mu=0 \quad \sigma=0.1$} &\multicolumn{2}{c}{$\mu=0 \quad \sigma=0.2$}  \\
    \cmidrule(lr){2-3} \cmidrule(lr){4-5}  
      & \multicolumn{1}{c}{mAP@.5} & \multicolumn{1}{c}{mAP@[.5, .95]} & \multicolumn{1}{c}{mAP@.5} & \multicolumn{1}{c}{mAP@[.5, .95]}  \\
     \midrule
     FRCNN (PAMI, '17) \cite{ren2015faster}  & 57.6 (+0.0) & 20.0 (+0.0) & 39.1 (+0.0) & 10.4 (+0.0) \\
     LIM (ICCV, '21)  \cite{tao2021towards}  & 72.6 \textcolor[rgb]{0.1961, 0.8039, 0.1961}{(+15.0)} & 28.1 \textcolor[rgb]{0.1961, 0.8039, 0.1961}{(+8.1)} & 60.4 \textcolor[rgb]{0.1961, 0.8039, 0.1961}{(+21.3)} & 19.3 \textcolor[rgb]{0.1961, 0.8039, 0.1961}{(+8.9)}\\
     SDANet (IJCV, '23) \cite{zhang2023pidray}  & 57.2 \textcolor{red}{(-0.4)} & 19.9 \textcolor{red}{(-0.1)} & 34.8 \textcolor{red}{(-4.3)} & 8.9 \textcolor{red}{(-1.5)} \\
     GADet (SENS J., '24)  \cite{10322651}  & 69.4 \textcolor[rgb]{0.1961, 0.8039, 0.1961}{(+11.8)} & 28.3 \textcolor[rgb]{0.1961, 0.8039, 0.1961}{(+8.3)} & 57.3 \textcolor[rgb]{0.1961, 0.8039, 0.1961}{(+18.2)} & 20.5 \textcolor[rgb]{0.1961, 0.8039, 0.1961}{(+10.1)} \\
     SCE (ICCV, '19) \cite{wang2019symmetric} & 53.1 \textcolor{red}{(-4.5)} & 18.4 \textcolor{red}{(-1.6)} & 58.7 \textcolor[rgb]{0.1961, 0.8039, 0.1961}{(+19.6)} & 21.0 \textcolor[rgb]{0.1961, 0.8039, 0.1961}{(+10.6)} \\
     LNCIS (ECCV, '20) \cite{yang2020learning} & 71.3 \textcolor[rgb]{0.1961, 0.8039, 0.1961}{(+13.7)} & 26.5 \textcolor[rgb]{0.1961, 0.8039, 0.1961}{(+6.5)} & 42.2 \textcolor[rgb]{0.1961, 0.8039, 0.1961}{(+3.1)} & 11.7 \textcolor[rgb]{0.1961, 0.8039, 0.1961}{(+1.3)}  \\
     OA-MIL (ECCV, '22) \cite{liu2022robust}  & 54.8 \textcolor{red}{(-2.8)} & 20.8 \textcolor[rgb]{0.1961, 0.8039, 0.1961}{(+0.8)} & 49.0 \textcolor[rgb]{0.1961, 0.8039, 0.1961}{(+9.9)} & 16.5 \textcolor[rgb]{0.1961, 0.8039, 0.1961}{(+6.1)}  \\
     {Mix-Paste+LLS} & \textbf{84.2 \textcolor[rgb]{0.1961, 0.8039, 0.1961}{(+26.6)}} & \textbf{35.8 \textcolor[rgb]{0.1961, 0.8039, 0.1961}{(+15.8)}} & \textbf{72.1 \textcolor[rgb]{0.1961, 0.8039, 0.1961}{(+33.0)}} & \textbf{25.7 \textcolor[rgb]{0.1961, 0.8039, 0.1961}{(+15.3)}} \\
    \bottomrule
  \end{tabular}}
  \label{tab:gaussian}
\end{table}

\noindent {
\textbf{Influence of Bounding Box Noise Distribution.}
Existing noisy-robust object detection methods (such as \cite{liu2022robust,li2020towards}) often apply the uniform distribution to generate bounding box noise. In this paper, we follow the same settings as these methods \cite{liu2022robust,li2020towards}. In this subsection, we also evaluate the effectiveness of our method by applying the
Gaussian distribution to generate bounding box noise. 
%We investigated the impact of bounding box noise distribution on the model's performance. 
Specifically, we apply  Gaussian noise to the training dataset and evaluate the detection performance. The shifting and scaling are changed as
%\begin{equation}
%   \begin{aligned}
   $\widetilde{x} = x + N(\mu, \sigma^2) \times w$, 
   $\widetilde{y} = y + N(\mu, \sigma^2) \times h$, 
   $\widetilde{w} = w \times (1 + N(\mu, \sigma^2))$,
    $\widetilde{h} = h \times (1 + N(\mu, \sigma^2))$,
%    \end{aligned}
%\end{equation}
where $N(\cdot)$ denotes the Gaussian distribution which is characterized by the mean ($\mu$) and the standard deviation ($\sigma$). {Note that in our experiments $\widetilde{x}, \widetilde{y}, \widetilde{w}$ and $\widetilde{h}$ are constrained to be positive numbers.}
The results are given in Table \ref{tab:gaussian}.}

{From Table \ref{tab:gaussian}, we can observe that 
our method 
can also effectively reduce the influence of bounding box noise generated by the  Gaussian distribution, showing the robustness of our method. }

\subsection{Visualization Results}\label{sec:vis}
{
We visualize some detection results obtained by the baseline method (i.e., FRCNN) and our method on the OPIXray dataset, as shown in Fig.~\ref{fig:results}. We can see that the baseline method gives false predictions with high confidence. 
In some cases,  both the predicted category labels and predicted bounding box coordinates are inaccurate (see the images in the first row). This is because the training of the baseline model is easily affected by noisy annotations. On the contrary, our method can give more accurate and correct predictions.
This indicates that our method is a simple yet effective data augmentation method, which can significantly improve the detection performance of X-ray prohibited items for learning with noisy annotations.}

\section{Conclusion}
\label{conclusion}
In this paper, we address the problem of training a robust X-ray prohibited item detector under noisy annotations from the novel perspective of data augmentation. We propose Mix-Paste by mixing multiple item patches with the same category label and generating a new image involving a mixed patch.  Such a manner not only effectively increases the probability of containing the correct prohibited item but also mimics item overlapping in X-ray images. 
{Moreover, we design an LSS strategy for loss calculation. Our strategy alleviates the negative influence of mistakenly treating potentially positive predictions as false predictions caused by the mixing process of item patches.}
We perform extensive experiments on two X-ray datasets to demonstrate the effectiveness of our method in training a noise-robust detector.
We also conduct experiments on the MS COCO dataset to verify the generalization ability of our method on the common object detection task.
These results 
demonstrate the benefits of data augmentation in tackling the challenges posed by learning with noisy annotations.


\bibliographystyle{IEEEtran}
\bibliography{reference}

% \newpage

\end{document}